%% file: ijcai23.tex
\documentclass{article}
\usepackage{ijcai23}

\usepackage{times}
\usepackage{soul}
\usepackage{url}
\usepackage[hidelinks]{hyperref}
\usepackage[utf8]{inputenc}
\usepackage[small]{caption}
\usepackage{graphicx}
\usepackage{amsmath}
\usepackage{amsthm}
\usepackage{booktabs}
\usepackage{algorithm}
\usepackage{algorithmic}
\usepackage{makecell}
\usepackage[switch]{lineno}
\newcommand{\nosection}[1]{\vspace{2pt}\noindent\textbf{#1.}}
\usepackage{tabularx}
\usepackage{multirow}
\usepackage{subfigure}
\usepackage{diagbox}
\hypersetup{
    colorlinks=true,
    linkcolor=blue,
    filecolor=blue,      
    urlcolor=blue,
    citecolor=cyan,
}

\title{SLPerf: a Unified Framework for Benchmarking Split Learning}

\author{
Tianchen Zhou\footnote{Equal contribution.}$^1$
\and
Zhanyi Hu\footnotemark[\value{footnote}]$^1$
\and
Bingzhe Wu$^2$\And
Cen Chen\footnote{{Corresponding Author.}}$^1$
\affiliations
$^1$School of data science \& engineering, East China Normal University, Shanghai, China\\
$^2$Tencent AI Lab, Shenzhen, China
\emails
\{tianchen, zhanyihu223\}@stu.ecnu.edu.cn,
bingzhewu@tencent.com,
cenchen@dase.ecnu.edu.cn
}

\begin{document}

\maketitle

\begin{abstract}
    Data privacy concerns has made centralized training of data, which is scattered across silos, infeasible, leading to the need for collaborative learning frameworks. To address that, two prominent frameworks emerged, i.e., federated learning (FL) and split learning (SL). While FL has established various benchmark frameworks and research libraries,SL currently lacks a unified library despite its diversity in terms of label sharing, model aggregation, and cut layer choice. This lack of standardization makes comparing SL paradigms difficult. To address this, we propose SLPerf, a unified research framework and open research library for SL, and conduct extensive experiments on four widely-used datasets under both IID and Non-IID data settings. Our contributions include a comprehensive survey of recently proposed SL paradigms, a detailed benchmark comparison of different SL paradigms in different situations, and rich engineering take-away messages and research insights for improving SL paradigms. SLPerf can facilitate SL algorithm development and fair performance comparisons. The code is available at \url{https://github.com/Rainysponge/Split-learning-Attacks}.
\end{abstract}

\section{Introduction}
Deep learning has strong application value and potential in various fields such as computer vision \cite{Krizhevsky2012ImageNet}, disease diagnosis \cite{8086133}, financial fraud detection \cite{DBLP:journals/corr/abs-2107-13673}, and malware detection \cite{DBLP:journals/corr/WangGZXGL16}. However, a large amount of high-quality labeled data is often required to train an effective deep-learning model \cite{CHERVENAK2000187}. Unfortunately, data are usually scattered across different silos or edge devices. Directly collecting them for training in a centralized fashion will inevitably introduce privacy issues. For instance, exchanging sensitive raw patient data between healthcare providers for machine learning could lead to serious privacy concerns \cite{HUANG2019103291,DBLP:journals/corr/LiuGNDKBVTNCHPS17}, which would no longer be permitted with the enforcement of data privacy laws \cite{Alessandro2013The}. Various collaborative learning methods have been proposed to facilitate joint model training without compromising data privacy \cite{LSDG2,DBLP:journals/corr/abs-1812-03288}, such as federated learning (FL) \cite{2016Communication} and split learning (SL) \cite{DBLP:journals/corr/abs-1810-06060,DBLP:journals/corr/abs-1912-12115}. FL enables collaboratively training a shared model while keeping all the raw data locally and exchanging only gradients. However, it often requires clients to have sufficient computational power for local training. To address the challenge of training heavy deep learning models on resource-constrained IoT devices, such as smartphones, wearables, or sensors, split learning is introduced. It splits the whole network into two parts that are computed by different participants (i.e., clients and a server), and necessary information is transmitted between them, including the data of the cut layer from clients and the gradients from the server.

Various benchmark frameworks or research libraries have been provided for FL such as FedML \cite{DBLP:journals/corr/abs-2007-13518}, PySyft\cite{DBLP:journals/corr/abs-1811-04017}, Flower\cite{FLOWER} and LEAF\cite{DBLP:journals/corr/abs-1812-01097}. However, unlike FL, there is no unified library for SL researchers yet. Therefore, despite significant progress made in SL research, there are several critical limitations that need to be addressed.

\begin{itemize}

\item  
\textbf{Lack of support for different SL paradigms}. 
SL has a high degree of diversity in terms of label sharing, model aggregation, etc. However, the lack of a comprehensive SL research library poses a problem for researchers to reinvent the wheel when comparing existing SL paradigms.

\item 
\textbf{Lack of a standardized SL evaluation benchmark}. 
The lack of a standardized SL evaluation benchmark makes it difficult to compare different paradigms fairly. With more paradigms being proposed, some studies train on CV datasets, while others train on sequential/time-series data \cite{DBLP:journals/corr/abs-2003-12365}. Dataset partitioning also varies, such as dividing based on labels or Dirichlet distribution in Non-IID settings. 

\end{itemize}


In this work, we systematically survey and compare different SL paradigms. 
We specifically classify these paradigms across various dimensions and present a unified research framework called SLPerf for benchmarking. To comprehensively verify the effectiveness of different SL paradigms, we conduct extensive experiments on four widely used datasets covering a wide range of tasks including computer vision, medical data analysis, tabular dataset, and drug discovery, under both IID and Non-IID data setting.  By analyzing these empirical results, we provide rich engineering take-away messages for guiding ML engineers to choose appropriate SL paradigms according to their specific applications. We also provide research insights for improving the robustness and efficiency of current SL paradigms.Compared to previous works, our study compares the performance of different SL paradigms with various data partitioning, providing a multi-dimensional perspective for evaluating SL paradigms. Our contributions are summarized as follows:
\begin{itemize}
\item \textbf{Unified framework for different SL paradigms.}
We introduce our framework SLPerf, an open research library and benchmark to facilitate SL algorithm development and fair performance. 
\item \textbf{Comprehensive survey of the recently proposed SL paradigms.}
We conduct a survey of recently proposed SL paradigms and classified them according to their characteristics. 
We also provide some suggestions for which scenarios are suitable for each SL paradigm.
\item \textbf{Detailed comparison of different SL paradigms.}
Prior research lacks detailed experimental comparisons of different SL paradigms. In this paper, we conduct experiments to compare different SL paradigms in different situations and provide a benchmark for SL. 
\item \textbf{Applying SL to graph learning.}
Sharing raw drug data is often not possible due to business concerns, making it challenging to build ML models. Our paper proposes using SL paradigms to solve this problem and has shown promising results on the Ogbg-molhiv dataset.

\end{itemize}
\begin{table*}[htbp]
\caption{The comparison of SL paradigms.}
\label{The comparison of Split learning paradigms.}
\begin{center}
\begin{small}
\begin{sc}
\begin{tabular}{ c|lcc}
\toprule
&paradigms &{\upshape Reduce Communication Cost}  & {\upshape Label Protection}\\
\midrule
\multirow{4}{*}{\upshape Model-Split-Only}&{\upshape Vanilla\cite{DBLP:journals/corr/abs-1810-06060}} & $\times$ & $\times$\\
&{\upshape U-shape \cite{DBLP:journals/corr/abs-1810-06060}}& $\times$ & $\surd$  \\
&{\upshape PSL\cite{2020Privacy}}     & $\times$& $\times$\\
&{\upshape AsyncSL\cite{2021Communication}}   & $\surd$& $\times$\\ \midrule
\multirow{6}{*}{\upshape Weight Aggregation-based}&{\upshape SplitFed\cite{SplitFed}} & $\times$ & $\times$\\
&FSL\cite{9582171} & $\surd$ & $\times$\\
&{\upshape FeSTA\cite{FeSVT}}   & $\times$& $\surd$\\
&{\upshape 3C-SL\cite{3C_SL}}& $\surd$ & $\times$ \\
&{\upshape HSFL\cite{857534b2836744119e61880381fa08e8_HFSL}}&$\surd$ & $\times$ \\
&{\upshape CPSL\cite{2022Split_CPSL}}&$\surd$ & $\times$ \\
\midrule
\multirow{3}{*}{\upshape Intermediate Data Aggregation-based}&{\upshape SGLR\cite{2021Server}}   & $\times$ &$\times$\\ 
&{\upshape LocFedMix-SL\cite{10.1145/3485447.3512153}}& $\times$& $\times$\\
&{\upshape CutMixSL \cite{10.1145/3485447.3512153}} & $\surd$& $\times$\\
\bottomrule
\end{tabular}
\end{sc}
\end{small}
\end{center}
\end{table*}

\begin{figure}[t!]
\centering
\includegraphics[width=0.4\textwidth]{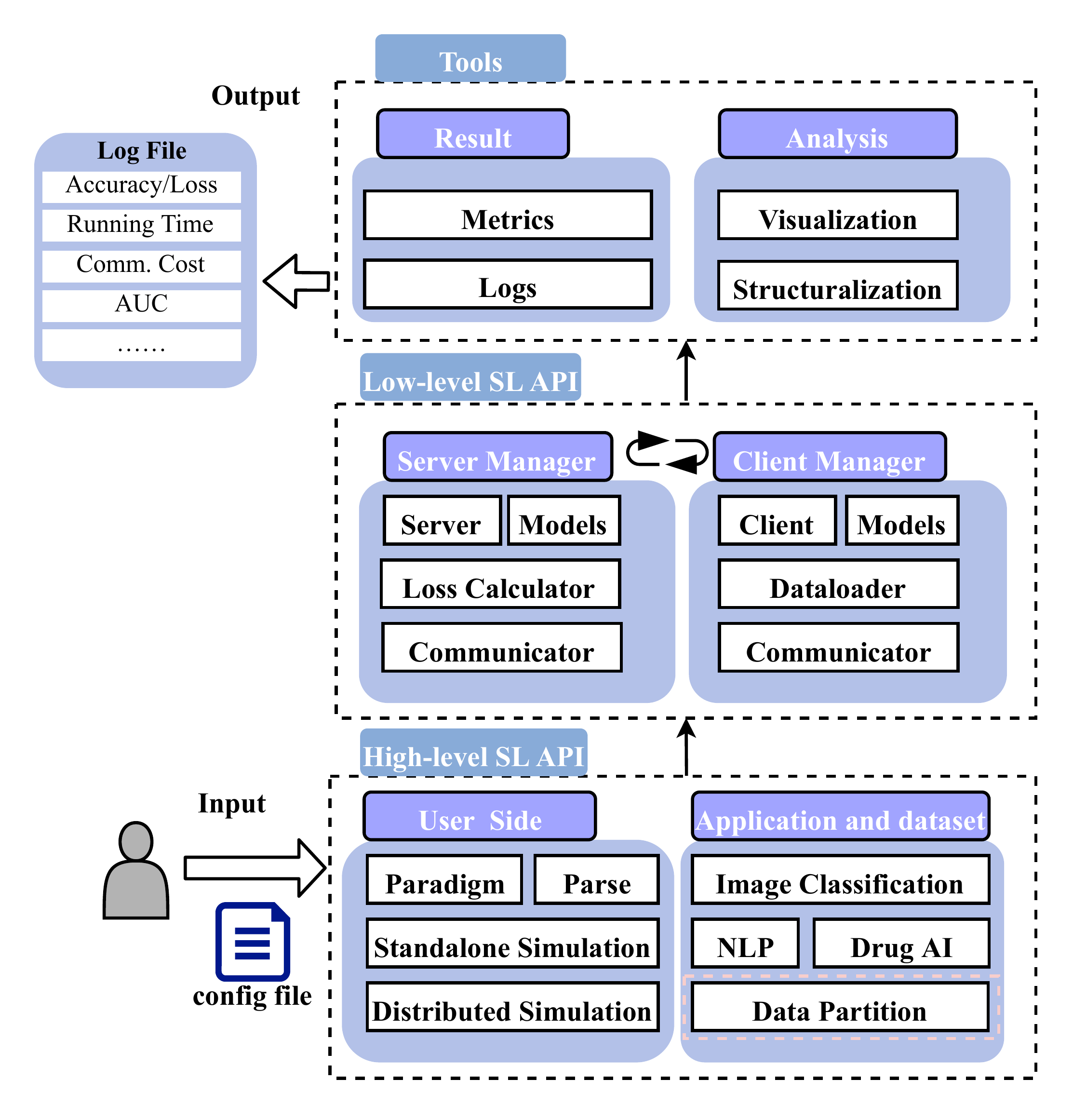}
\caption{Overview of our proposed framework SLPerf.}
\label{fig:framework}
\end{figure}
\section{Framework}
We propose SLPerf, an open research library for SL that offers various interfaces and methods to rapidly develop SL algorithms while enabling fair performance comparisons across different configurations. SLPerf supports various SL training configurations, including dataset partitioning methods, numbers of clients, and paradigms. It also provides standardized benchmarks with explicit evaluation metrics to validate baseline results, facilitating fair performance comparisons. Moreover, our framework includes standardized implementations of multiple SL paradigms, allowing users to familiarize themselves with our framework's API.

Figure~\ref{fig:framework} provides an overview of our framework. We divide the framework into three modules: High-level SL API, Low-level SL API, and Tools. When conducting SL experiments, users will use the functions from High-level SL API by providing a configuration file that includes the selection of the paradigm, dataset, and data partitioning method. Then, using the corresponding factory, functions from Low-level SL API will be called to generate  clients and server objects for training. Once the training is completed, the experimental results will be recorded in a log file that contains details such as accuracy, AUC, communication cost, and other pertinent information. Users can visualize and analyze the data using the code provided in the Tool Module. If users need to develop their own SL paradigms, they can utilize the interfaces provided in the Low-level SL API to custimize. By overloading certain methods, such as determining whether to aggregate model weights, they can define the necessary communication information.

\input{Paradigm}
\input{eval}

\section{Conclusion}
In this paper, we propose SLPerf, an open research library and benchmark designed for SL researchers to facilitate the development of SL paradigms. We summarize and categorize existing SL paradigms and compare their performance on four widely-used datasets under both IID and Non-IID data settings. Our experiments demonstrate the importance of investigating how SL can perform better under Non-IID data settings and determining the optimal layer for model splitting.

\appendix
\section*{Acknowledgments}
This work was supported by the National Natural Science Foundation of China under Grant No. 62202170.




\bibliographystyle{named}
\bibliography{ijcai23}

\end{document}

%% file: Paradigm.tex
\section{Paradigm}
\label{Paradigm section}



In this section, we will present an overview of various split learning (SL) paradigms. We will conduct a comprehensive comparison of these paradigms using Table ~\ref{The comparison of Split learning paradigms.}, which lists mainstream SL paradigms and compares them based on various features. 
We categorize paradigms into three groups based on whether intermediate data or model weights are aggregated during training. 
The “model-split-only” group partitions the model architecture across participants without data or weight aggregation. The “weight aggregation-based” group uses techniques like FedAvg \cite{2016Communication} to aggregate model weights from different devices. The “intermediate data aggregation-based” group aggregates intermediate data like gradients or smashed data. This categorization provides a clear understanding of SL approaches for researchers and practitioners.

\subsection{Model-Split-Only}
\begin{figure}[htbp]
\centering

\includegraphics[width=0.8\linewidth]{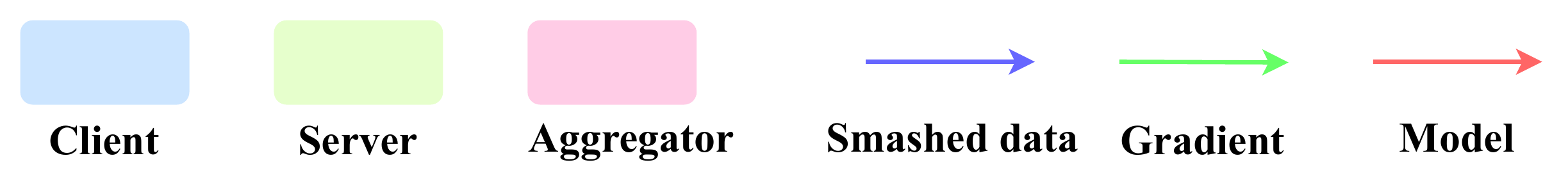}

\subfigure[]{
    \centering
    \includegraphics[width=0.2\textwidth]{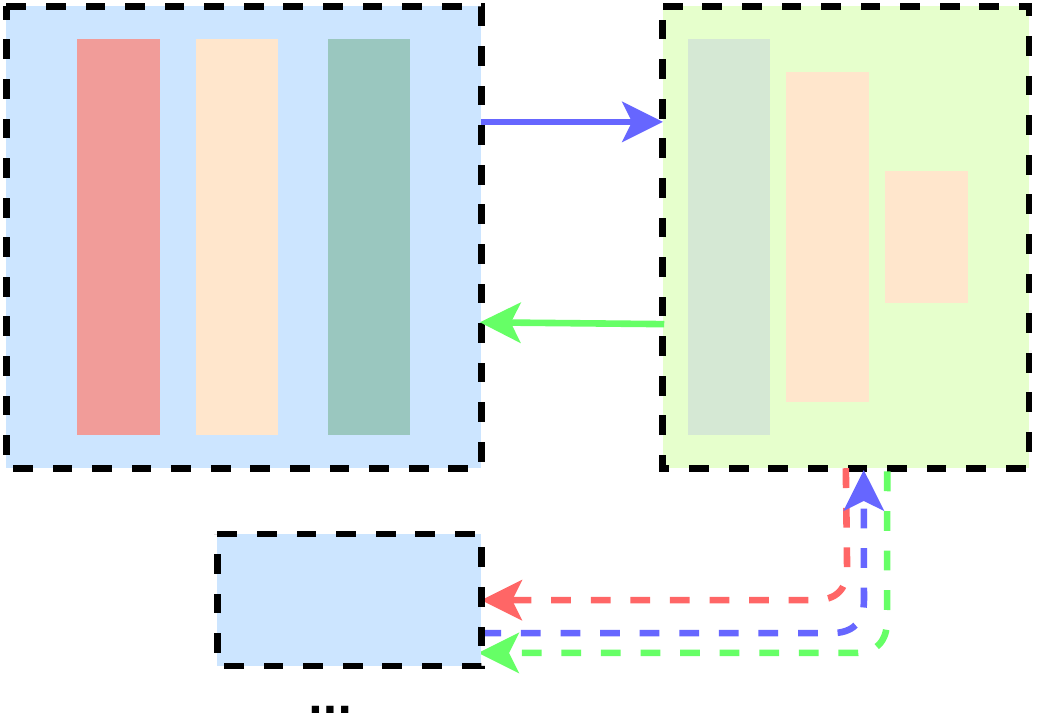}
    \label{fig:VanillaSL: VanillaSL}
}
\subfigure[]{
    \centering
    \includegraphics[width=0.2\textwidth]{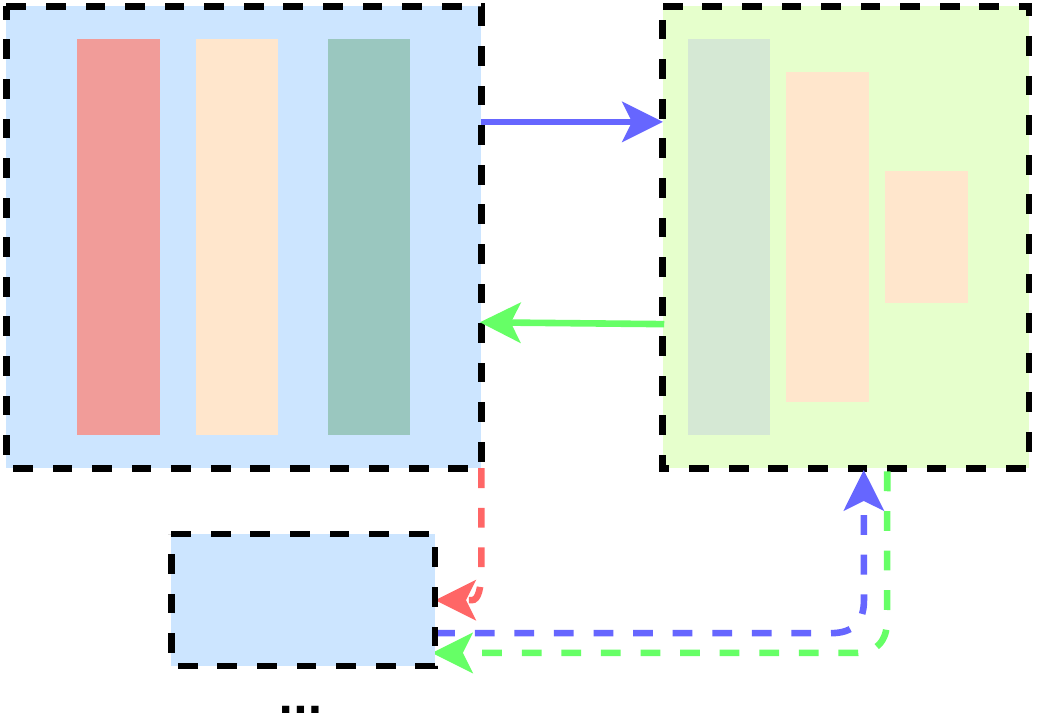}
    \label{fig:VanillaSL: VanillaSL2}
}
\caption{The overview of the Vanilla SL paradigms: (a) Vanilla (centralized) (b) Vanilla (peer-to-peer) }
\label{fig:VanillaSL}
\end{figure}

The Vanilla Split Learning paradigm \cite{DBLP:journals/corr/abs-1810-06060} divides the original model into two parts: the server model and the client model. The client trains the model locally using its own data and then transmits the “Smashed Data” to the server for updating the server model. The server sends the backward-propagated gradients back to the client for updating the client model. And each client must wait for the previous client to complete a training round before interacting with the server. This paradigm has two setups: (1) centralized, where the client uploads model-weight-containing files to a server or trusted third party, and the next client downloads the weights and trains the model, as shown in Figure~\ref{fig:VanillaSL: VanillaSL}; (2) peer-to-peer (P2P), where the server sends the previous trained client’s address to the current client for transferring the weights, as shown in Figure~\ref{fig:VanillaSL: VanillaSL2}.
However, the Vanilla paradigm has a label leakage issue which is addressed by the U-shape paradigm \cite{DBLP:journals/corr/abs-1810-06060}. The U-shape paradigm divides the original model into three parts: the head and tail on the client-side and the body on the server-side. The client only sends smashed data to the server, which forwards it for forward propagation and sends computed results back to the client. The tail calculates loss and gradient, updates with backpropagation, and sends the gradient to the server. The server updates the body and sends the backpropagation result to the client for updating the head.

\begin{figure}[htbp]
\centering
\includegraphics[width=0.8\linewidth]{pic/Lag.pdf}

\subfigure[]{
\includegraphics[width=0.22\textwidth]{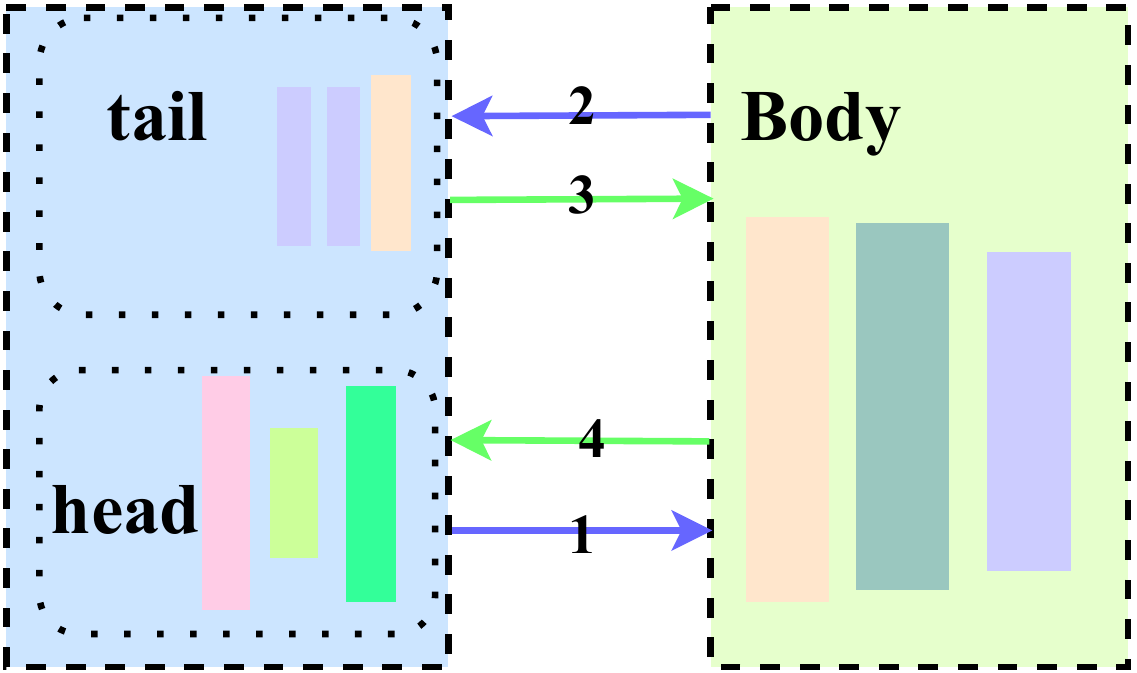}
\label{fig:USHAPE: USHAPE_SL}
}
\subfigure[]{
\includegraphics[width=0.15\textwidth]{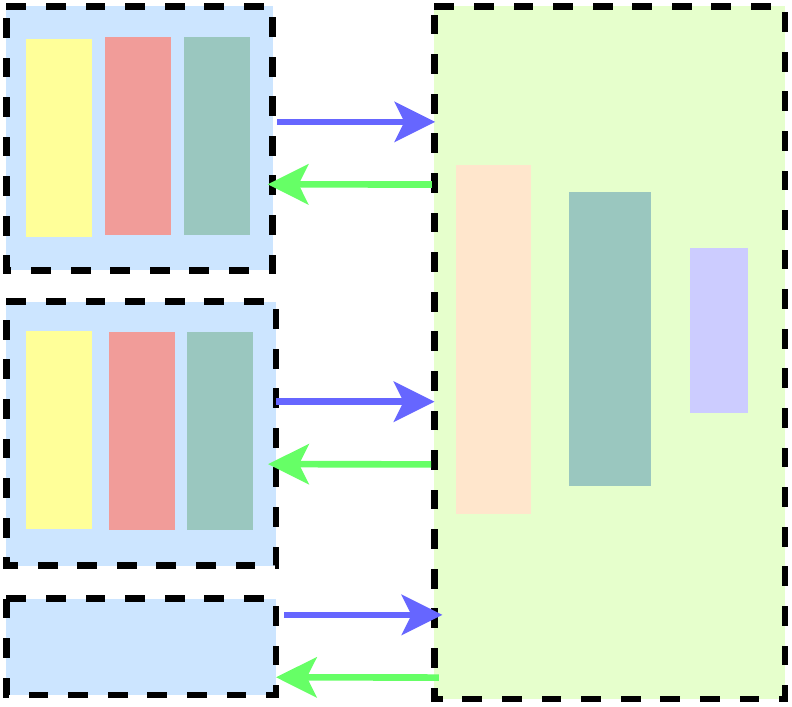}
\label{fig:parallelSL: parallelSL}
}
\caption{The overview of Model-Split-Only paradigms: (a) U-shape (b) PSL}
\label{fig:USHAPE}
\end{figure}
The synchronous training properties of Vanilla SL may result in significant delays, which can be mitigated through parallel training. To address this issue, the Parallel Split Learning (PSL) \cite{2020Privacy} paradigm has been proposed, as shown in Figure~\ref{fig:parallelSL: parallelSL}. In PSL, clients initiate forward propagation in parallel and transmit the smashed data $z_i$ (where $i$ denotes the index of the client node) to the server. Upon completion of forward and backward propagation, the server sends back the computed gradient $g_i$ derived from $z_i$ to the i-th client, reducing the overall training time.


For reducing the communication cost, the Split Learning using Asynchronous Training (AsyncSL) \cite{2021Communication} paradigm has been developed, which is a loss-based asynchronous training scheme that updates the client-side model less frequently and only sends/receives smashed data /gradients in selected epochs. 
When the state is A, clients send smashed data to the server, and receive the gradient from the server. In state B, only the client sends smashed data to the server, and the server does not send the gradient back. State C involves no exchange of information between clients and the server. Also to reduce the communication cost, the circular convolution-based batch-wise compression for SL (C3-SL) \cite{DBLP:journals/corr/abs-2207-12397} is a paradigm combining circular convolution with SL. Clients use cyclic convolution \cite{Plate1995Holographic} to compress multiple features into a single compressed feature, which is decoded on the server side through cyclic correlation \cite{Plate1995Holographic}


\begin{figure}[htbp]
\centering
\includegraphics[width=0.9\linewidth]{pic/Lag.pdf}

\subfigure[]{
\includegraphics[width=0.45\linewidth]{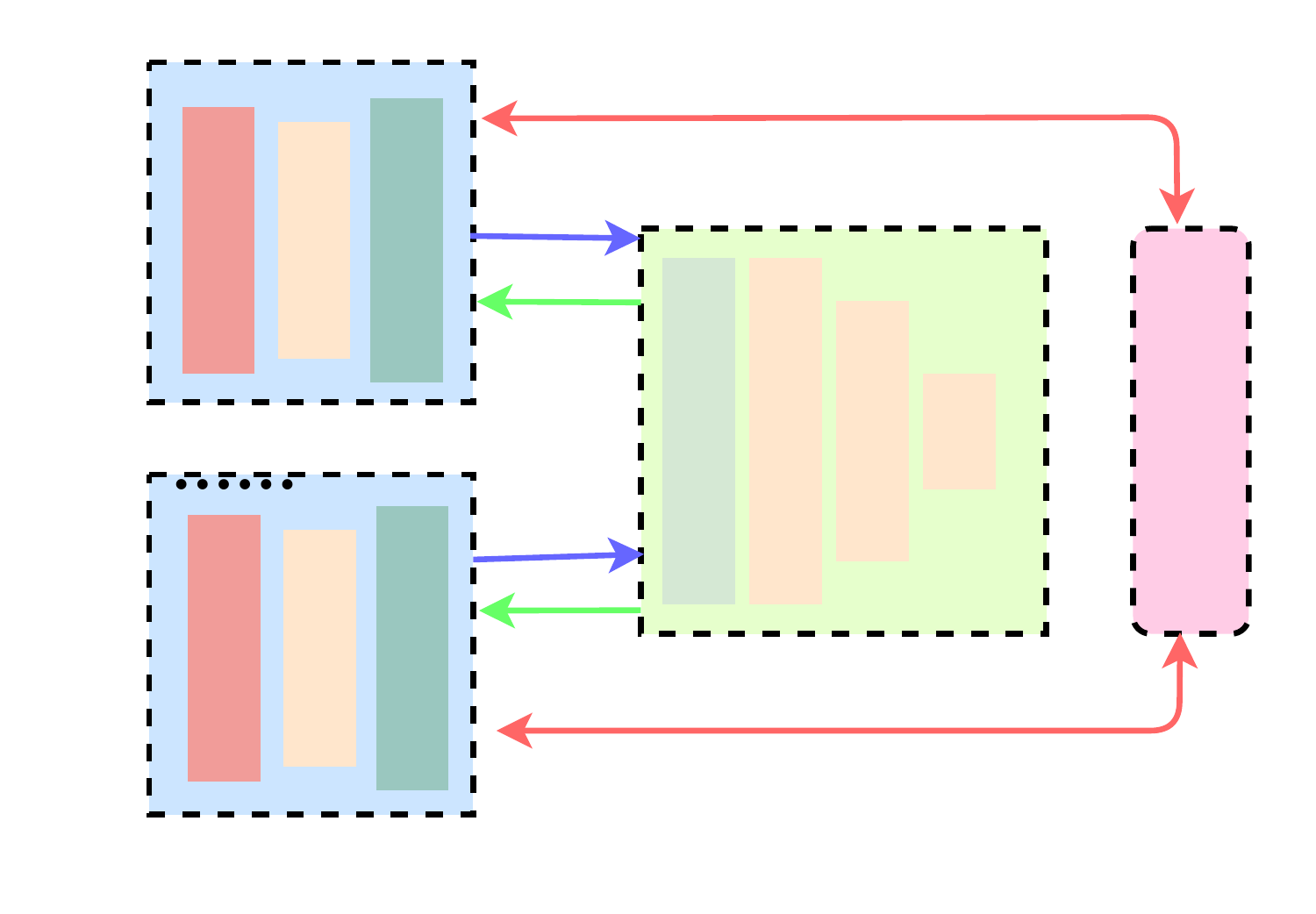}
\label{fig:AVERAGE: splitFed}
}
\subfigure[]{
\includegraphics[width=0.22\textwidth]{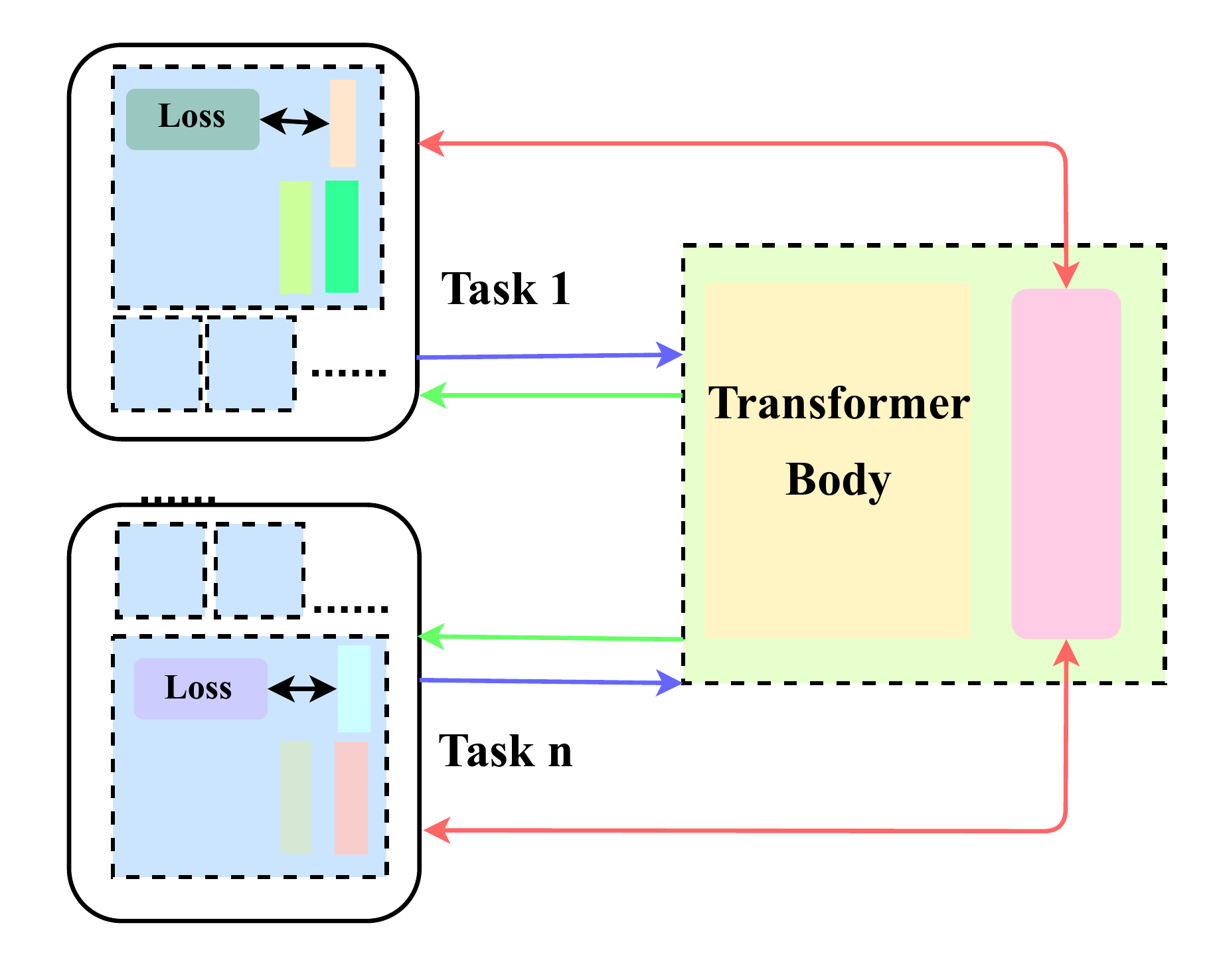}
\label{fig:USHAPE: multiTask}
}
\caption{The overview of Weight Aggregation-based SL paradigms: (a) SplitFed (b) FeSTA.}
\label{fig:AVERAGE}
\end{figure}
\subsection{Weight Aggregation-based}
Weight Aggregation-based SL paradigms use Weight Aggregation to train a shared client model by aggregating locally trained models from clients. During the training process, some or all of the clients will send their models to the server or a trusted third party for model weight aggregation at a certain stage, such as SplitFed\cite{SplitFed}, FSL\cite{9582171}, HSFL\cite{857534b2836744119e61880381fa08e8_HFSL}. Weight Aggregation-based SL can handle resource-limited clients such as those found in IoT and is advantageous for addressing Non-i.i.d. data scenarios.

In SplitFed\cite{SplitFed}, the model is split into two parts and trained in a similar way to PSL. And after each client completes one round of training, its local model weights are transmitted to the server, which aggregates the local models using the FedAvg aggregation method. The process is illustrated in Figure~\ref{fig:AVERAGE: splitFed}. Additionally, the authors propose a variant, SplitFedv2\cite{SplitFed}, in which clients are trained linearly in order, rather than in parallel. In the context of Federated Split Learning (FSL) \cite{9582171}, each client trains its local model in accordance with the SL paradigm in conjunction with its corresponding Edge Server, instead of interacting with a central server. Subsequent to the completion of one epoch, the model weights of the Edge Server are aggregated via the central server.

FeSTA \cite{FeSVT} is a paradigm that utilizes the Vision Transformer and the concept of U-shape paradigms. In this paradigm, clients train various models and are divided into groups depending on their tasks and trained in parallel. After each training epoch, clients send the model head and tail parameters to the server, which aggregates the model parameters for the same task and returns them to the clients using techniques such as FedAvg. This paradigm is suitable for training models for multiple similar tasks as it has a shared Transformer body.

Hybrid Split and Federated Learning (HSFL) \cite{857534b2836744119e61880381fa08e8_HFSL} and Cluster-based Parallel SL (CPSL) \cite{2022Split_CPSL} are paradigms designed to address the practical constraints of power, bandwidth, and channel quality typically present in client devices.  Both methods utilize greedy algorithms to optimize their problems.
\begin{figure}[htbp]
\centering
\includegraphics[width=0.9\linewidth]{pic/Lag.pdf}

\quad
\subfigure[]{
\includegraphics[width=0.45\linewidth]{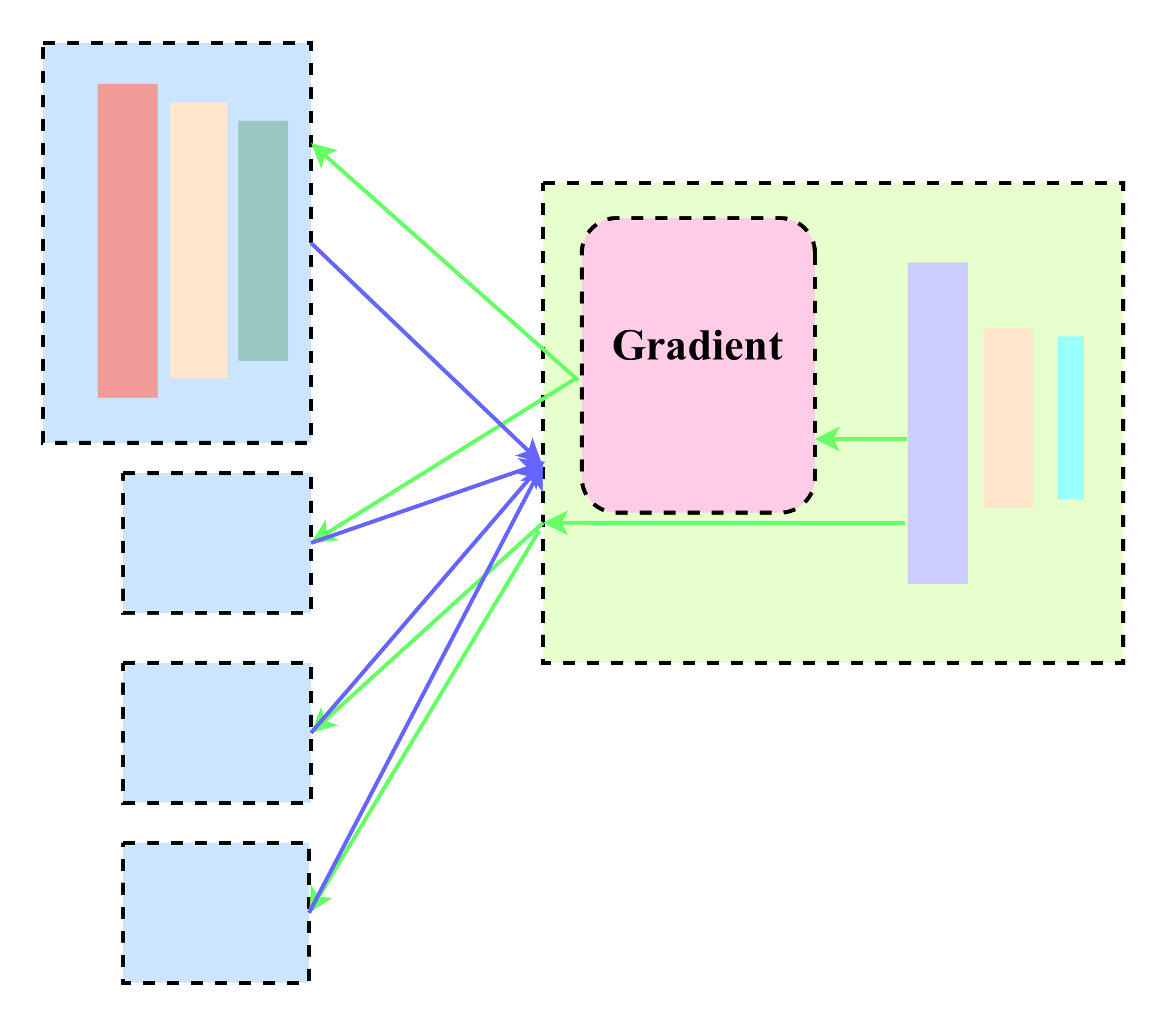}
\label{fig:AVERAGE: SLGR}
}
\subfigure[]{
\includegraphics[width=0.22\textwidth]{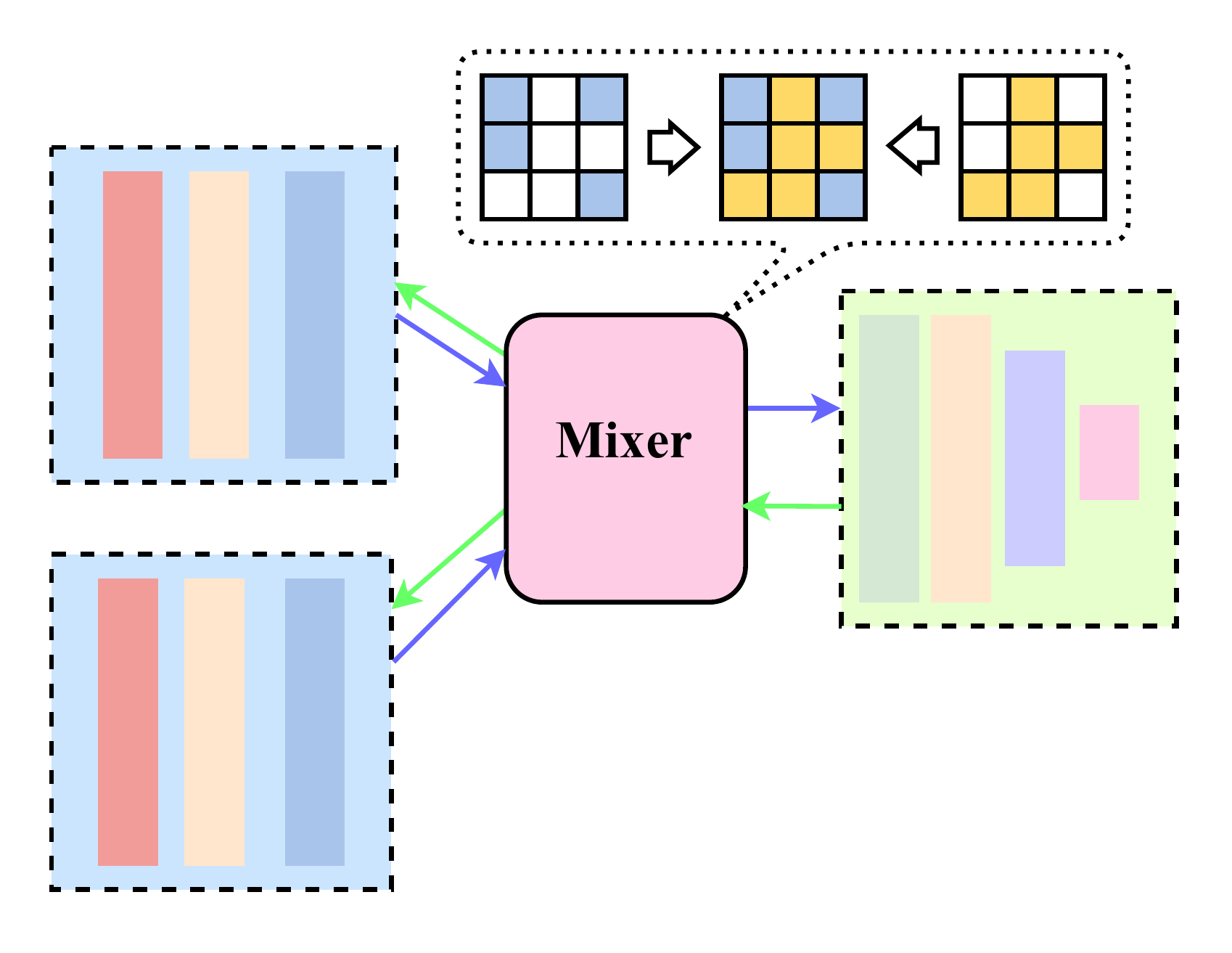}
\label{fig:communication_cost: MIXer}

}

\caption{The overview of Intermediate Data Aggregation-based SL paradigms: (a) SLGR (b) CutMixSL.}
\label{fig:communication_cost}
\end{figure}
\subsection{Intermediate Data Aggregation-based}
Unlike model weight aggregation, some SL paradigms aggregate intermediate data, such as gradients and smashed data, from different clients to enable the exchange of local data information and minimize communication costs.

In SGLR \cite{2021Server}, the server aggregates the local gradients of a subset of clients rather than the model parameters. This approach reduces the amount of information shared among clients and eliminates the client decoupling problem. The learning rate is also separated into two parts to improve the overall training performance and address the Server-Side Large Effective Batch Problem. The process is shown in Figure~\ref{fig:AVERAGE: SLGR}. Also, inspired by existing learning rate acceleration work \cite{DBLP:journals/corr/GoyalDGNWKTJH17}, the learning rate is separated into two parts. The server model’s learning rate is accelerated to solve the Server-Side Large Effective Batch Problem \cite{2021Server}.

In LocFedMix-SL\cite{10.1145/3485447.3512153}, the server mixes up the smashed data from different clients to create new smashed data for forward propagation. This smashed data mixing increases the effective batch size at the model's top layers to match its higher effective learning rate. Each client has an auxiliary network designed to maximize the mutual information between the smashed data and input data which can make the relationship between the smashed data and model input more tight, thereby improving model performance and training speed. In CutMixSL \cite{2022Visual}, the authors combined SL with ViT and proposed a new type of data called CutSmashed. As shown in Figure~\ref{fig:communication_cost: MIXer}, each client creates their own CutSmashed data by randomly masking a portion of their original smashed data. Then, clients send this partial data to a Mixer, which combines them and sends them to the server for training. This method reduces communication costs as clients only need to send a portion of their data, and the server can use multi-casting to send back the gradient. And CutSmashed data increases data privacy as adversaries find it difficult to reconstruct the untransmitted regions.

%% file: eval.tex
\section{Evaluation}
\begin{table*}[htbp]
\caption{The overview of different SL paradigms on different settings. Centralized learning results are provided in bold.}
\label{summary list}
\begin{center}
\begin{small}
\begin{sc}
\resizebox{\textwidth}{13mm}{
\begin{tabular}{c|ccc|ccc|ccc|cc}
\toprule
\multirow{2}{*}{\diagbox{paradigm}{$\alpha$}} 
 &  \multicolumn{3}{c}{MNIST(Acc = \textbf{0.9843})}  & \multicolumn{3}{c}{CIFAR-10(Acc = \textbf{0.9114})}& \multicolumn{3}{c}{Adult(Acc = \textbf{0.8173})}& \multicolumn{2}{c}{molhiv(AUC = \textbf{0.7472})} \\
\cline{2-12} &
  0 & 0.10&
   \makecell[c]{$+\infty$ \\ (IID)}
  & $0$ & $0.10$ & \makecell[c]{$+\infty$ (IID)}& $0$ & 0.1 & \makecell[c]{$+\infty$ (IID)} & \makecell[c]{$+\infty$ (IID)} & 0.10\\
\midrule
{\upshape SplitFed}    & 0.9082 & 0.9768 & 0.9882 &0.6155&0.7781&0.9089 &0.7217&0.7832&0.8027&0.7374&0.6524\\
{\upshape PSL}    & 0.3486& 0.6764&  0.9713 &0.4023&0.6484&0.9139 & 0.5319 &0.6042&0.8163&0.7341&   0.7053   \\
{\upshape Vanilla}     & 0.1826&0.7288& 0.9792 & 0.3424&0.3534&0.9008 & 0.5171&0.5612&0.7979&0.7428&0.7004\\
{\upshape U-shape} & 0.3297&0.7019&0.9728& 0.3691& 0.3513& 0.9067 & 0.5239 & 0.5957&0.8078&-&-\\
{\upshape SGLR} & 0.3312&0.8603&0.9734&0.4588&0.6765&0.8954& 0.5953 &0.6101&0.7932&-&-\\
\bottomrule
\end{tabular}
}
\end{sc}
\end{small}
\end{center}
\end{table*}
As the above sections show, our framework has implemented the current mainstream
SL paradigms and support typical data splitting schemes including IID and Non-IID.
In this part, we provide comprehensive empirical studies of these paradigms against a wide range of tasks including computer vision, medical data analysis, tabular dataset, and drug discovery. By summarizing and analyzing these empirical results, we provide rich engineering take-away messages for guiding the ML engineer to choose appropriate SL paradigms according to their own applications. We also provide research insights for improving current SL paradigms from the dimension of robustness, efficiency, and privacy protection. As a summary, we first introduce evaluation setup in Section \ref{sec:Evaluation setup}. 
Then we demonstrate empirical experimental results and provide detailed analysis in Section \ref{sec:Basic analysis}.

\subsection{Evaluation setup}
\label{sec:Evaluation setup}
We introduce the basic evaluation setup used in this paper including dataset/model setup, evaluation criteria, and run-time environment.
\subsubsection{Dataset and model setup}
To evaluate our framework, we conducted extensive experiments using various models and datasets. For all experiments, we trained the models using SGD with a fixed learning rate of 0.01 on minibatches of size 64.
We evaluated different SL paradigms on four commonly-used datasets, namely, MNIST, CIFAR-10, UCI Adult, and Ogbg-molhiv, which cover a wide range of data types, including images, tabular data, and graphs.

\nosection{MNIST} 
The MNIST dataset \cite{1990Handwritten} is a collection of handwritten digits from postal codes, with 70,000 samples divided into 60,000 training and 10,000 testing samples. The digits are centered on a 28x28 grid, and the goal is to classify them into 0-9. We trained the LeNet model on this dataset and split the network layers after the 2D MaxPool layer.

\nosection{CIFAR-10} 
The CIFAR-10 dataset comprises 60,000 32x32 RGB color images with 10 exclusive class labels, divided into 50,000 training samples and 10,000 testing samples. The task is to classify objects in the images.
ResNet56 \cite{ResNet} is used for evaluation, with the model split at the first ReLU activation layer.

\nosection{Ogbg-molhiv} 
The ogbg-molhiv dataset contains molecular samples and labels indicating their ability to inhibit HIV replication. This dataset is used to model a graph learning task, where the molecular data is represented as a graph with atoms as nodes and edges between them. 
This task is an example of molecular property prediction, which is crucial in the AI-drug domain.


\nosection{Adult Data Set}
The UCI Adult  dataset~\cite{osti_421279} is a classic data mining benchmark with 48,842 instances from UC Irvine repository \footnote{{\tt
http://archive.ics.uci.edu\\/ml/datasets/Census+Income
}}. The dataset has 14 features such as country, age, education. The goal for this binary classification task is to predict whether income exceeds $50K/yr$ based on the census information. We use a network consists of a fully connected (FC) layer, a ReLu activation layer and an other FC layer which is split at the first ReLu activation layer.


\subsection{Analysis}
\label{sec:Basic analysis}
In this section, we demonstrate our empirical results and try to analyze them from the following dimensions:
\begin{itemize}
    \item Comparison among different SL paradigms mentioned in Table~\ref{The comparison of Split learning paradigms.}. We provide the results of different SL paradigms when training on IID and Non-IID settings.
    \item  Comparison with typical FL paradigm FedAvg \cite{2016Communication}.  
    \item The comparison of the communication costs of different SL paradigms.
    \item Comparison with on-device local training. 
\end{itemize}

\begin{figure*}[t]
\centering
\includegraphics[width=0.6\textwidth]{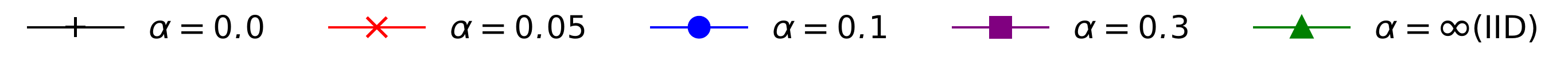}

\subfigure[]{
\includegraphics[width=0.2\textwidth]{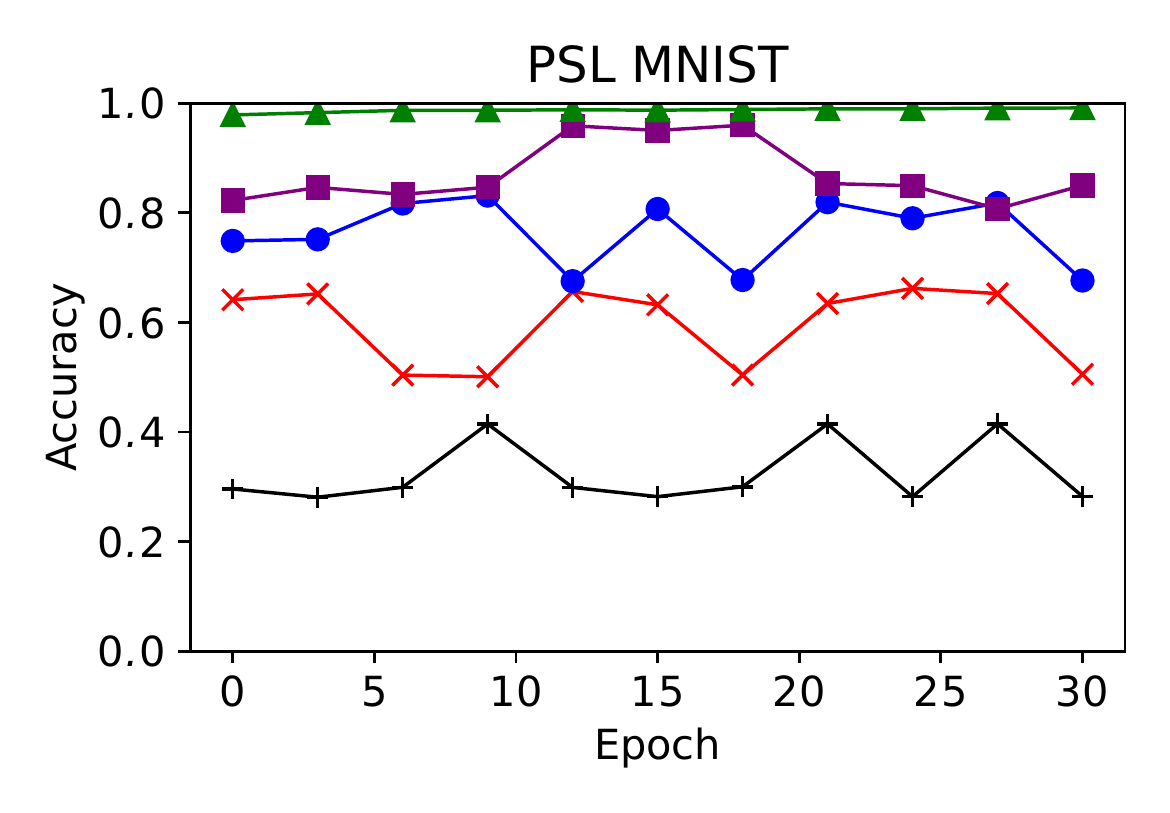}
\label{fig:Dirichlet MNIST:PSL}
}
\subfigure[]{
\includegraphics[width=0.2\textwidth]{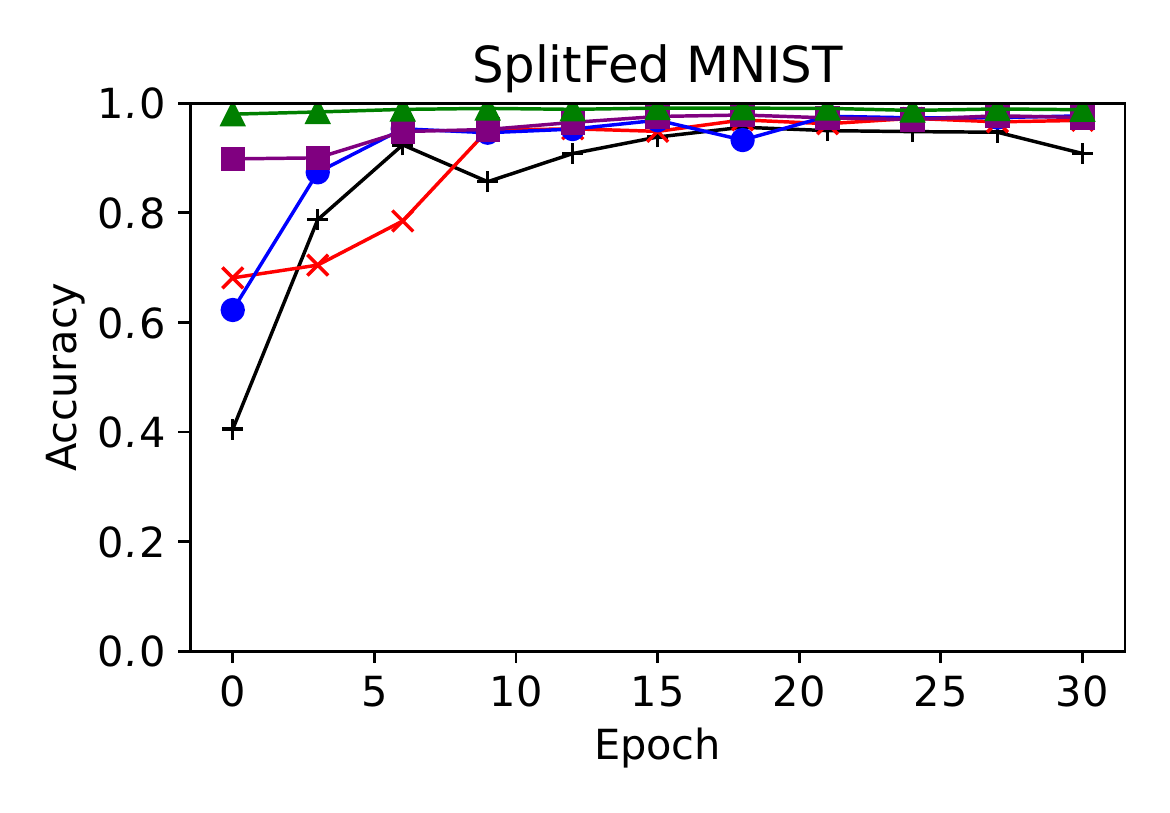}
\label{fig:Dirichlet MNIST:SFL}
}
\subfigure[]{
\includegraphics[width=0.2\textwidth]{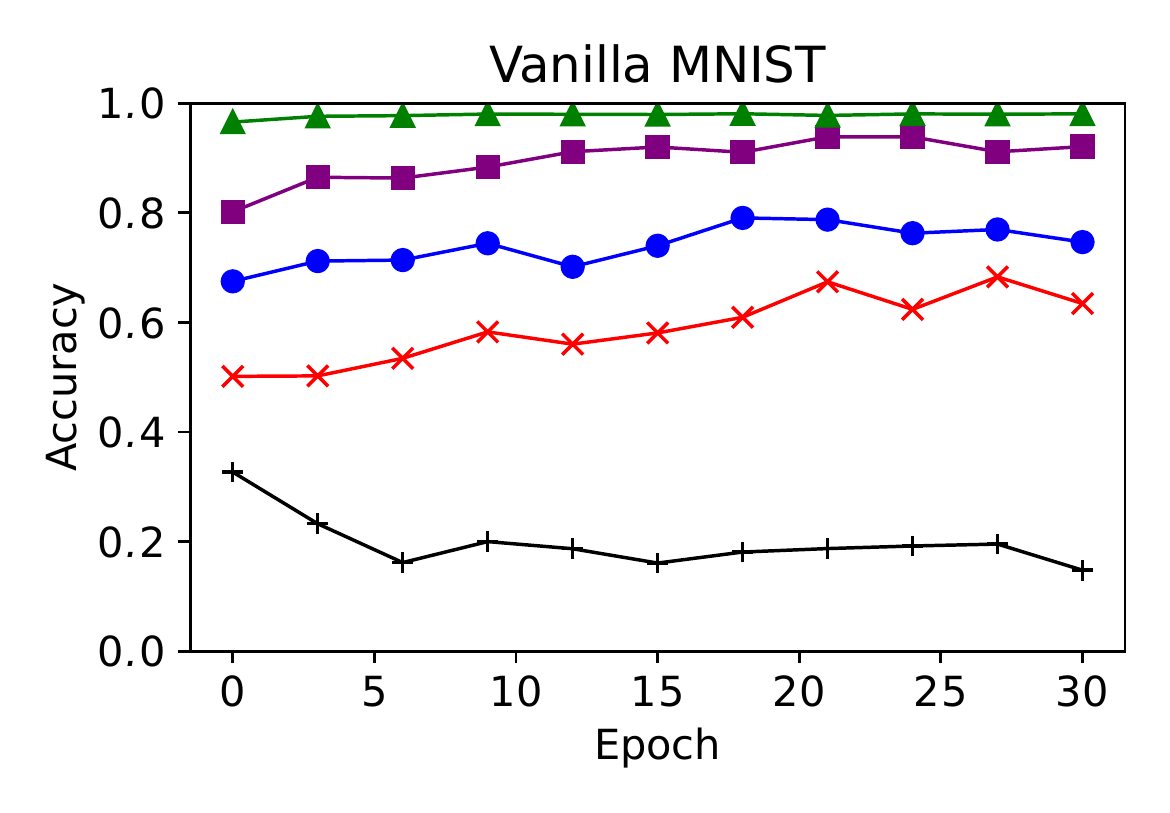}
\label{fig:Dirichlet MNIST:Vanilla}
}
\subfigure[]{
\includegraphics[width=0.2\textwidth]{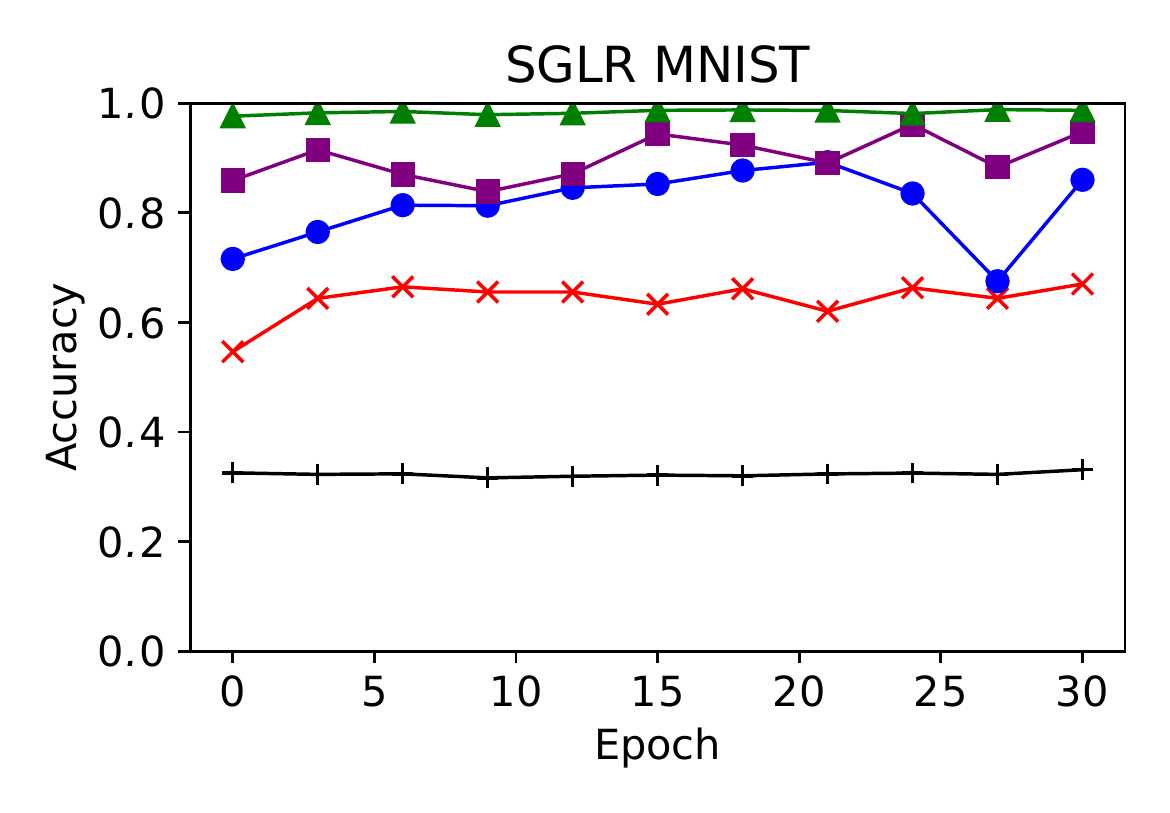}
\label{fig:Dirichlet MNIST:SGLR}
}
\subfigure[]{
\includegraphics[width=0.2\textwidth]{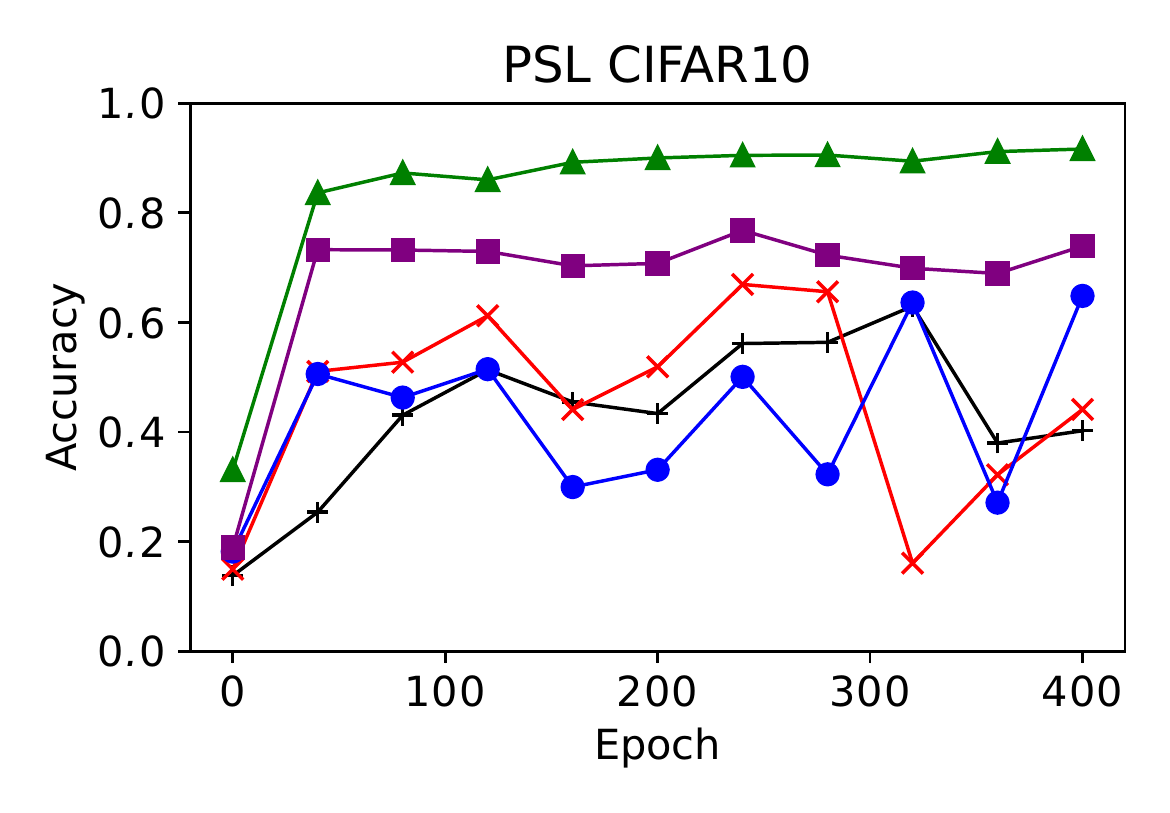}

\label{fig:Dirichlet cifar:PSL}
}
\subfigure[]{
\includegraphics[width=0.2\textwidth]{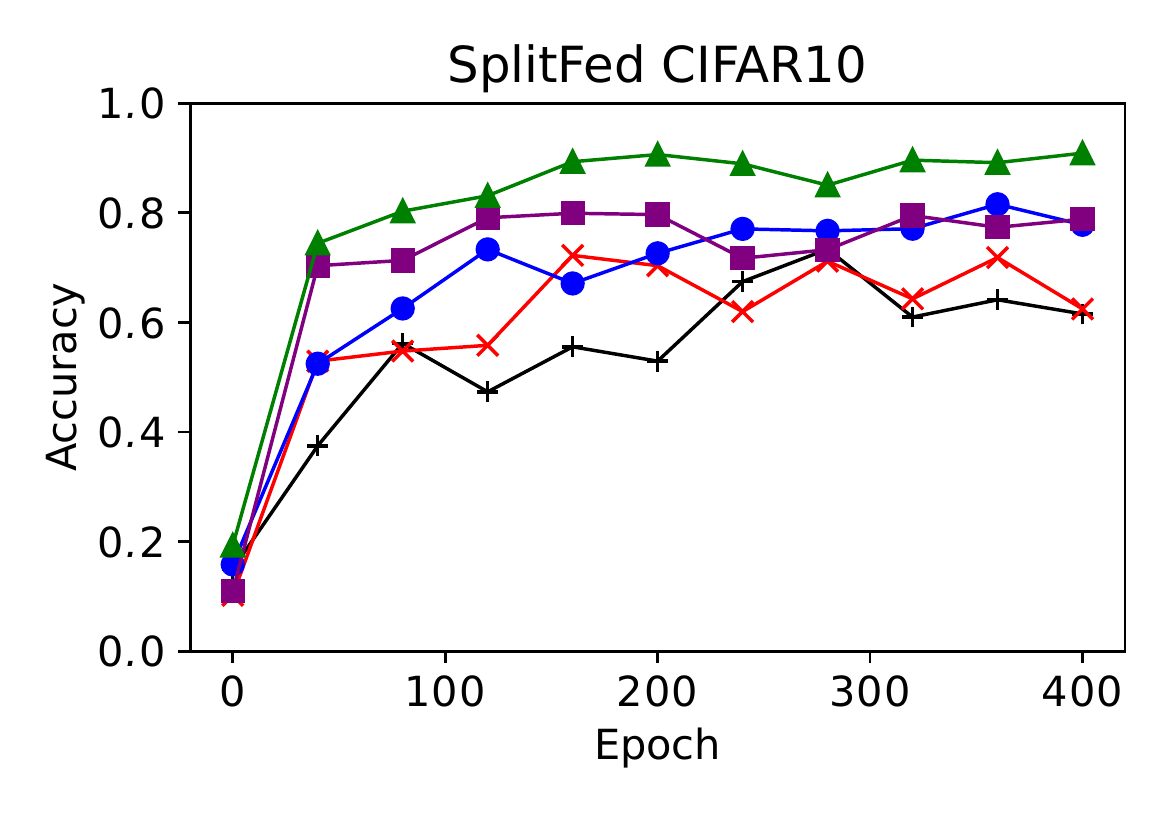}
\label{fig:Dirichlet cifar:SFL}
}
\subfigure[]{
\includegraphics[width=0.2\textwidth]{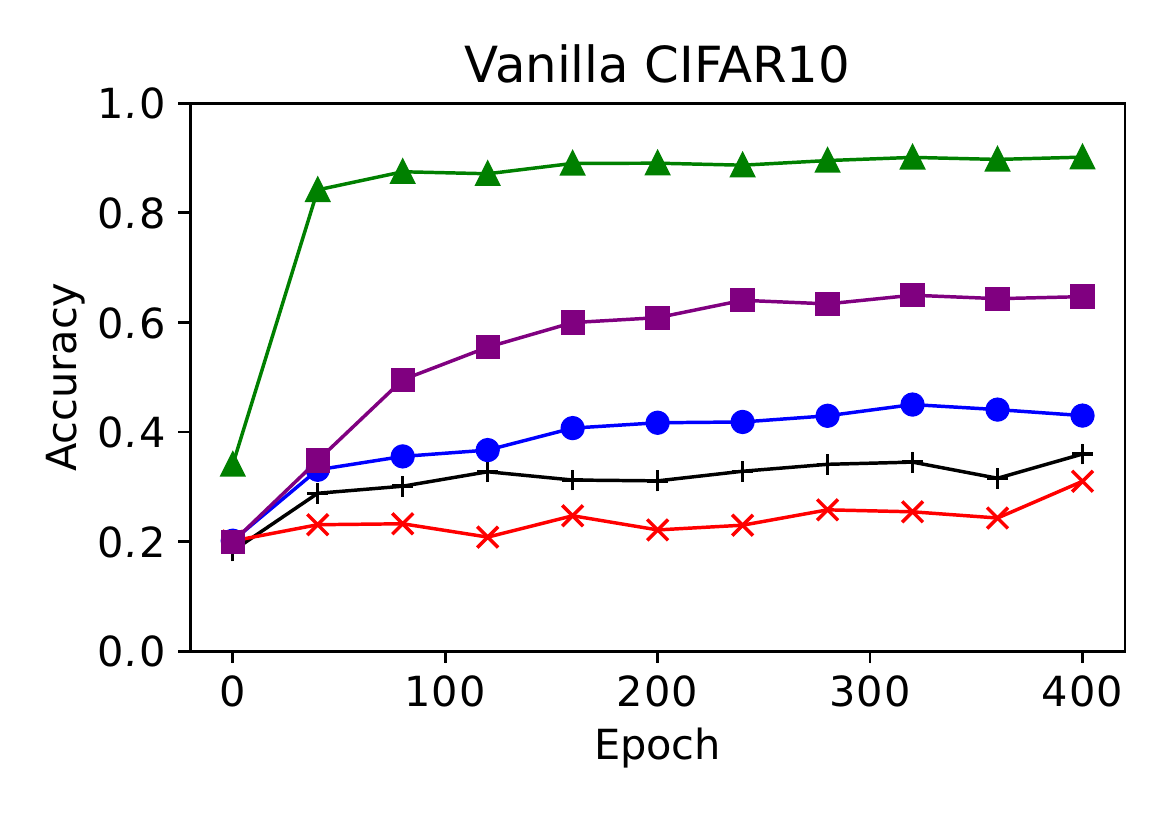}
\label{fig:Dirichlet cifar:Vanilla}
}
\subfigure[]{
\includegraphics[width=0.2\textwidth]{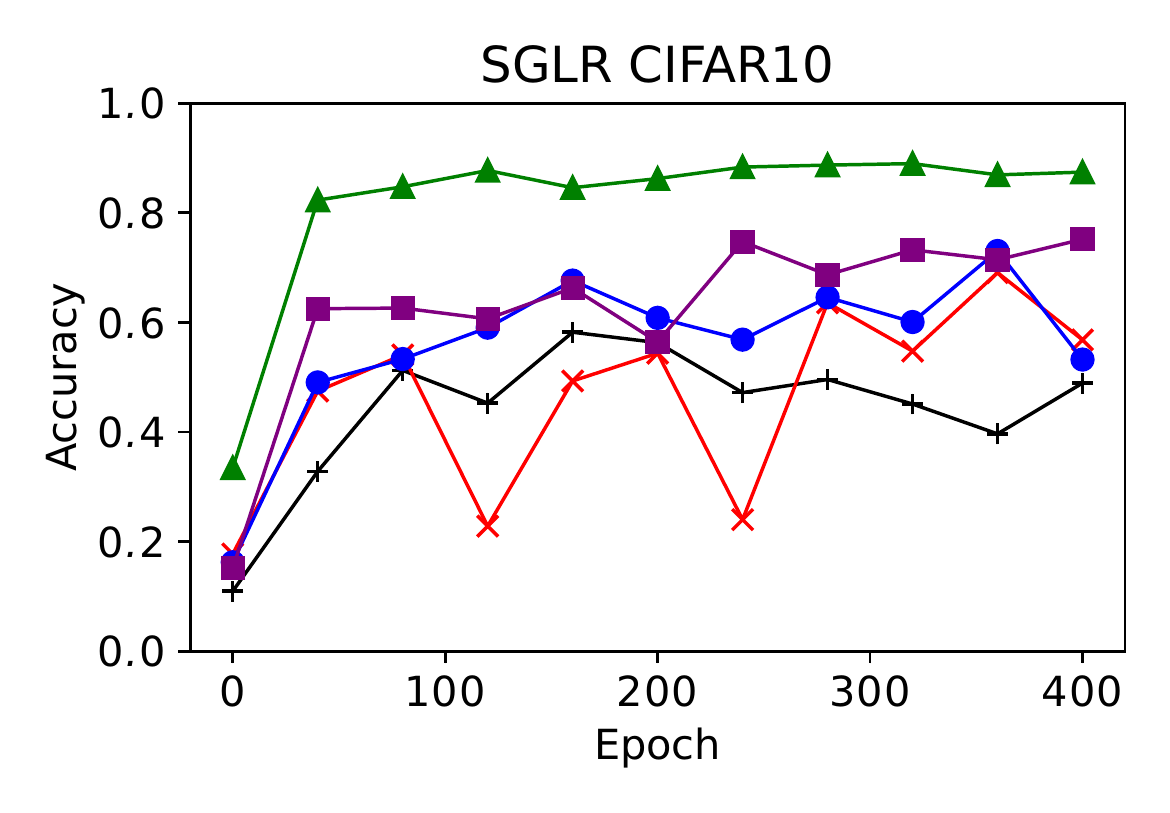}
\label{fig:Dirichlet cifar:SGLR}
}
\caption{The performance of three SL paradigms on different $\alpha$ when the number of clients is 3 on the MNIST and CIFAR-10 datasets.}
\label{fig:Dirichlet MNIST}
\end{figure*}

\subsubsection{Comparisons in the IID setting}
\label{sec: analysis of IID}
First, we investigate the performance of different SL paradigms in the IID setting. The overall results are shown in Table ~\ref{summary list}. For most of the datasets we studied in this paper, all SL paradigms have achieved similar accuracy and are comparable with the centralized one. For example, on the CIFAR-10 dataset with 3 clients, the Vanilla SL paradigm has achieved an accuracy of 90.08$\%$ while the centralized training achieves 91.14$\%$. In contrast, a commonly-used paradigm named U-shape has achieved the accuracy of centralized training. 
For the graph dataset, SL also have achieved a promising result, for example, vanilla achieved an AUC of 0.7428.

\subsubsection{Comparisons in the Non-IID setting}
\label{sec: analysis of Non IID}
In practice, the Non-IID setting is more frequently encountered than the IID setting.
To simulate such Non-IID settings, we employ label-based splitting for the MNIST and CIFAR-10 datasets. For UCI-Adult dataset, we use attribute-based splitting based on relationship. 
Following prior work, we first sample $p_{i,k}$
from Dirichlet distribution Dir($\alpha$) and then assign $p_{i,k}$ proportion of the samples of class/attribute $k$ to client $i$, where we set $\alpha$ from 0.05 to 0.3 to adjust the level of data heterogeneity
in our experiments. The smaller the $\alpha$, the larger the data heterogeneity.
When $\alpha = 0$, samples from different clients have non-overlap label/attribute spaces.  
For example, the MNIST dataset with 10 labels will be divided into five clients in this way, and each client will be assigned samples corresponding to two labels.
And We will follow the evaluation settings in Section \ref{sec:Evaluation setup} to discuss the performance of paradigms on Non-IID data. \\

The overall results are shown in Table ~\ref{summary list}. In this table, we select two representative values of $\alpha$ to demonstrate results of different paradigms on different datasets (i.e., $\alpha=0$ and $\alpha=0.1$ ). The client number is set to $3$ for all results in this table (More results with other client number are shown in the appendix). We summarize the following observations from the results:
\begin{itemize}

    \item 
    SL paradigms struggle to generalize to Non-IID settings, especially when data heterogeneity is high (small $\alpha$). Table~\ref{summary list} shows that most paradigms perform poorly on the CIFAR-10 when $\alpha=0$, with Vanilla achieving only 34.24$\%$ accuracy. When samples are not strictly divided by labels/attributes (e.g., $\alpha=0.1$), most SL paradigms' accuracy are still worse than centralized learning.
    
    \item 
    SplitFed has demonstrated higher robustness and higher performance compared to other methods. Figure~\ref{fig:Dirichlet MNIST:SFL} shows that SplitFed's performance remains stable when $\alpha$ is small(e.g., $\alpha=0.10$), achieving an accuracy of $97.68\%$. However, Vanilla and PSL's accuracy is low when $\alpha$ is small, as shown in Figure~\ref{fig:Dirichlet MNIST:Vanilla} and Figure~\ref{fig:Dirichlet MNIST:PSL}. Similar findings are shown in Figure~\ref{fig:Dirichlet cifar:PSL}-~\ref{fig:Dirichlet cifar:SGLR} on the CIFAR-10 dataset. SplitFed allows clients to share information about their model, while SGLR shares information by averaging the local gradients of the cut layer, achieving better performance than other paradigms except for SplitFed when $\alpha$ is small, such as an accuracy of $67.65\%$ when $\alpha=0.1$.
      
    \item Vanilla performs worse than other paradigms and even worse than local training in some cases. As shown in Table \ref{summary list}, Vanilla only achieves $18.26\%$ accuracy when $\alpha$ is 0. Vanilla's training pattern prevents the model from converging in these case.
    
\end{itemize}

\begin{figure}[t!]
\centering
\includegraphics[width=0.45\textwidth]{pic/legend2.pdf}
\subfigure[]{
\includegraphics[width=0.2\textwidth]{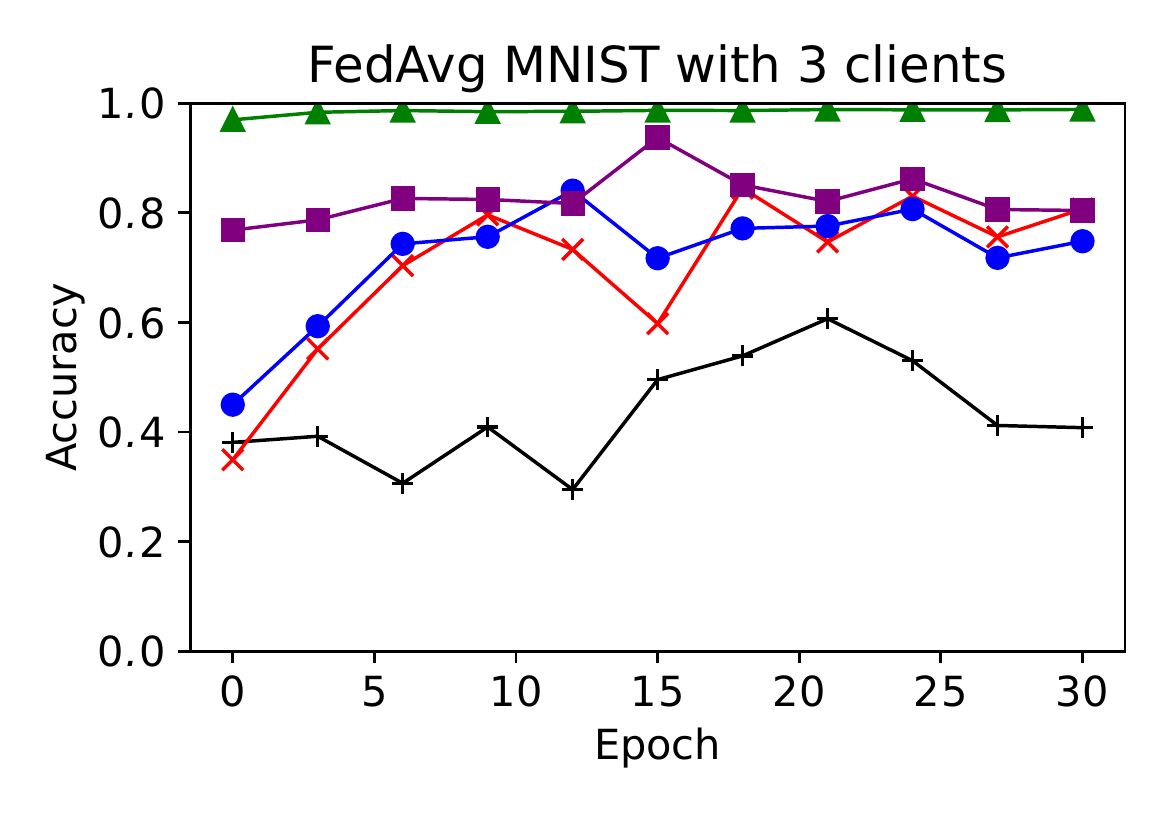}
\label{fig:fedavg Dirichlet:MNIST}
}
\subfigure[]{
\includegraphics[width=0.2\textwidth]{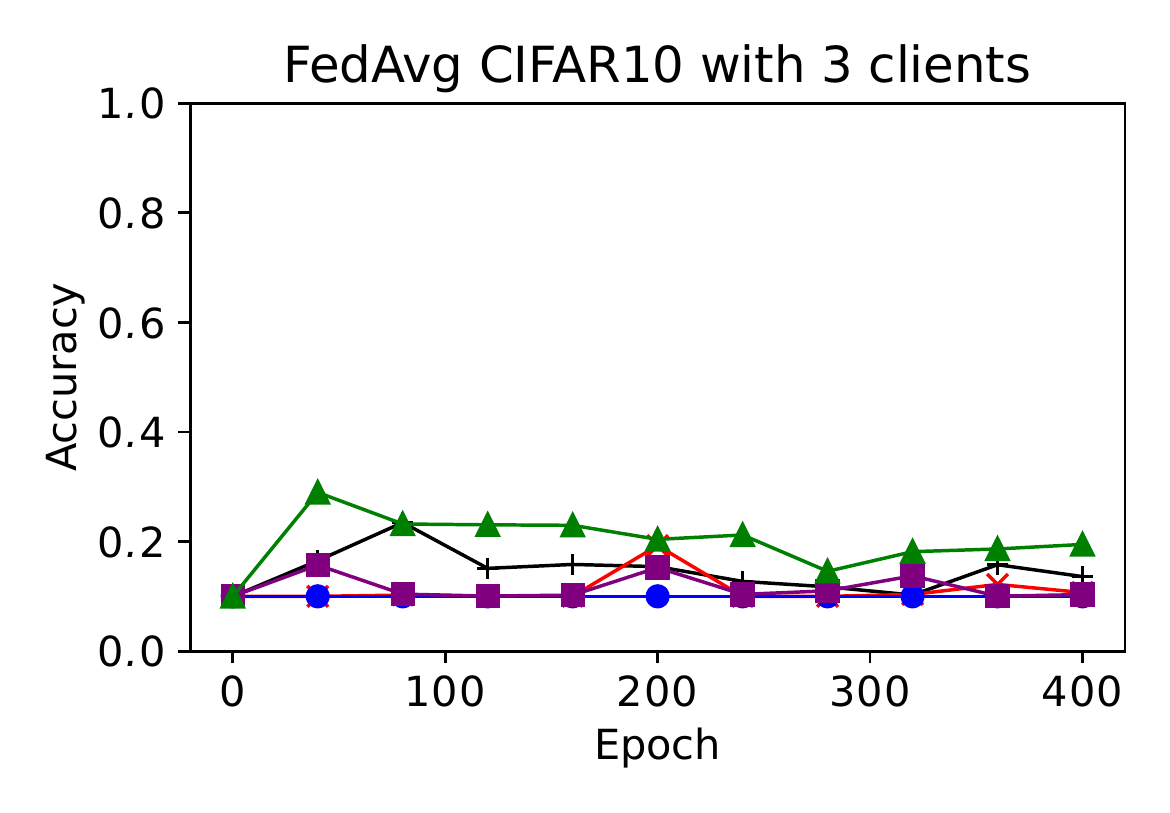}
\label{fig:fedavg Dirichlet:CIFAR10}
}
\subfigure[]{
\includegraphics[width=0.2\textwidth]{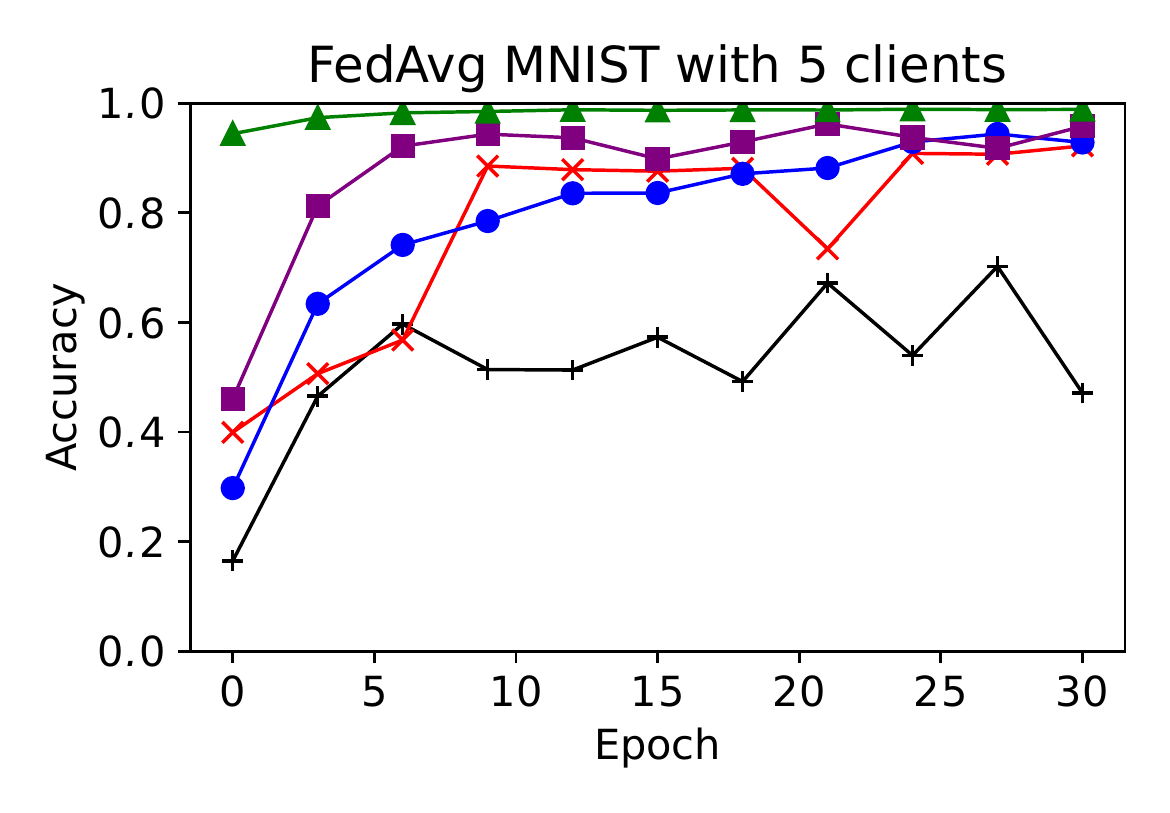}
\label{fig:fedavg Dirichlet:MNIST_5}
}
\subfigure[]{
\includegraphics[width=0.2\textwidth]{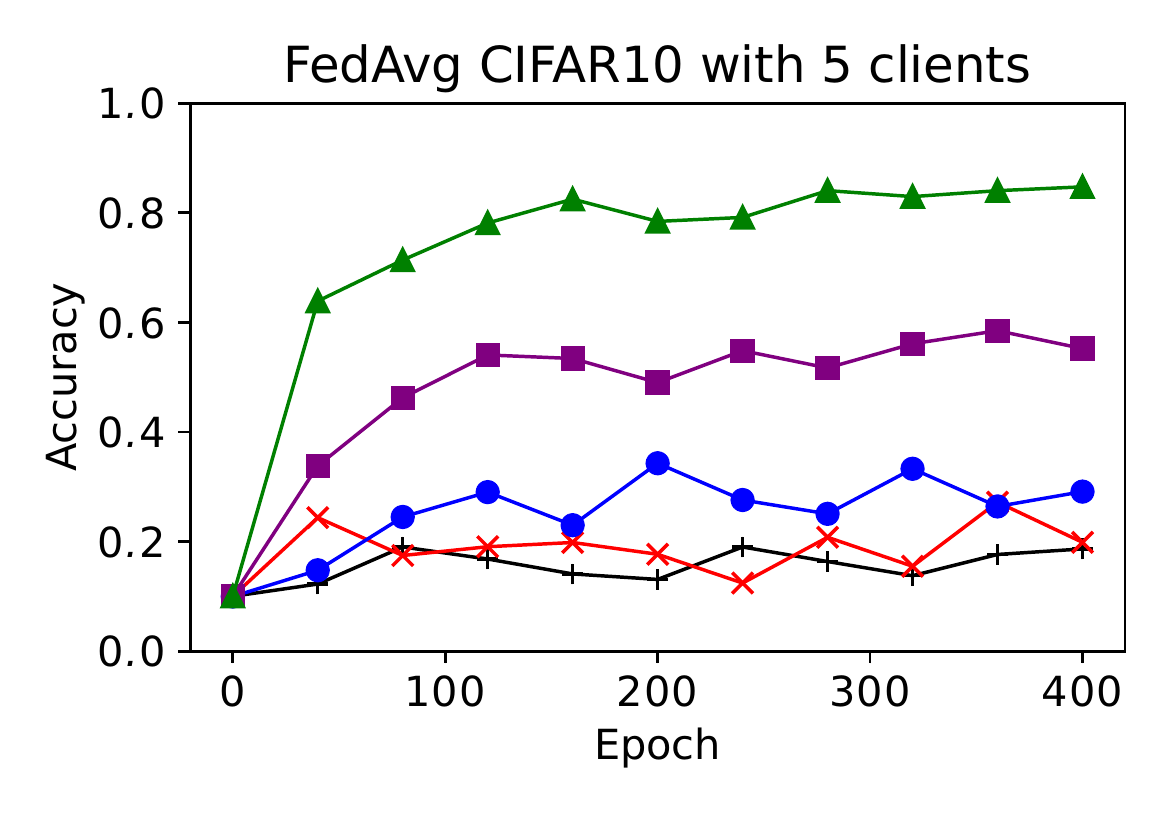}
\label{fig:fedavg Dirichlet:CIFAR10_5}
}
\caption{The performance of FedAvg with different $\alpha$ on the MNIST dataset and CIFAR-10 dataset. In (a)(b) we set the number of clients as 3, and in (c)(d) the number of clients is 5.}
\label{fig:fedavg Dirichlet}
\end{figure}


\subsubsection{Comparison with typical FL paradigm}

\label{sec: compare with FL}

In this section, we compare the performance of FedAvg with that of SL paradigms on the MNIST and CIFAR-10 datasets. The experimental results of SL paradigms are shown in Figure~\ref{fig:Dirichlet MNIST}, and those of FedAvg are shown in Figure~\ref{fig:fedavg Dirichlet}.

First, we observe that on both two datasets SplitFed can always converge faster than FedAvg with the same data partition method, as shown in Figure~\ref{fig:Dirichlet cifar:SFL} and Figure~\ref{fig:fedavg Dirichlet:CIFAR10} .


Second, Fedavg performs better than other SL paradigms when the dataset and model are simple on Non-IID setting, such as the MNIST dataset and LeNet-5. For example, as illustrated in Figure \ref{fig:Dirichlet MNIST:PSL} -\ref{fig:Dirichlet MNIST:SGLR} and Figure \ref{fig:fedavg Dirichlet:MNIST}, when $\alpha$ is small, FedAvg has shown superior accuracy compared to most SL paradigms, with the exception of SplitFed. 

However, when the data and models become more complex, FedAvg learns much more slowly than SL paradigms. For instance, on the CIFAR-10 dataset, most SL paradigms can achieve better accuracy than FedAvg, as shown in Figure~\ref{fig:fedavg Dirichlet:CIFAR10} and ~\ref{fig:Dirichlet cifar:PSL}-~\ref{fig:Dirichlet cifar:SGLR}. 

\subsubsection{The communication cost of different paradigms}
\label{sec: communication cost}


We compared the communication cost of PSL, SplitFed, and AsyncSL, representing intermediate data aggregation, model aggregation, and reduced communication respectively. The communication cost is computed by the size of messages transferred between clients and the server. The $loss_{threshold}$ \cite{2021Communication} we set in AsyncSL is 0.10. We split the networks at different layers to compare the impact of the choice of the cut layer. In Table~\ref{communication cost MNIST}, the $small$ indicates a network with one convolution-layer and $large$ indicates a network with two convolution-layers.

\begin{table}[t]
\caption{The communication cost of training per epoch on MINST when the number of clients is 3.}
\label{communication cost MNIST}
\vskip 1pt
\begin{center}
\begin{small}
\begin{sc}
\begin{tabular}{ccccc}
\toprule
\multirow{2}{*} {paradigm} &  \multicolumn{2}{c}{small(MB)} & \multicolumn{2}{c}{large(MB)} \\
\cline{2-3} \cline{4-5} 
  &send & receive&send & receive\\
\makecell[c]{\upshape SplitFed \\ \upshape half data}    & \makecell[c]{18.33\\ 9.36}& \makecell[c]{197.59\\ 98.86} & \makecell[c]{4.04\\3.97} &\makecell[c]{58.70\\29.40} \\
\cline{1-5}
\makecell[c]{\upshape PSL \\ \upshape half data}  & \makecell[c]{ 17.86\\8.79 } &\makecell[c]{197.59\\ 98.89 } &  \makecell[c]{3.93\\ 3.88 } & \makecell[c]{58.65\\29.42}   \\
\cline{1-5}
\makecell[c]{\upshape AsyncSL\\ \upshape  half data }     & \makecell[c]{4.58\\ 2.43 } &\makecell[c]{28.24\\ 15.43}& \makecell[c]{1.84\\ 1.03} & \makecell[c]{8.39\\ 4.32}\\
\bottomrule
\end{tabular}
\end{sc}
\end{small}
\end{center}
\end{table}





\begin{itemize}


    \item Dataset size affects communication cost. Table ~\ref{communication cost MNIST} shows that communication cost decreases significantly when only half of the dataset is used.
    \item Different SL paradigms result in different communication costs. For example, SplitFed sends more data than PSL because clients in SplitFed need to transmit local model parameters to the server. 
    However, when the model is more complex or the number of clients is much larger, the communication cost of models cannot be ignored \cite{DBLP:journals/corr/abs-1812-03288}.
    \item The choice of the cut layer may have an impact on the communication cost. Table ~\ref{communication cost MNIST} shows that when training on the MNIST dataset, communication cost decreases significantly when clients hold two convolution layers instead of one. This is because the choice of the cut layer affects the size of the smash data and gradients.

\end{itemize}

\subsubsection{Comparison with on-device local training}
\label{Local training section}
One of the primary objectives of SL is to enhance the model's performance of each client without requiring the exchange of raw data. Therefore, this section aims to compare the performance of SL with on-device local training. Thus, we include an extra set of experiments to demonstrate the effectiveness of on-device local training. In these experiments, each client follows the same data partitioning process as in the other SL experiments and employs an incomplete dataset to train their local model without any communication. 

\begin{figure}[t]
  \centering
  \includegraphics[width=0.46\textwidth]{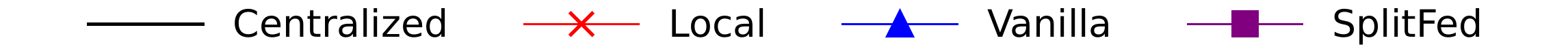}
  \\
  \subfigure[The accuracy against the client number]{
  \includegraphics[width=0.215\textwidth]{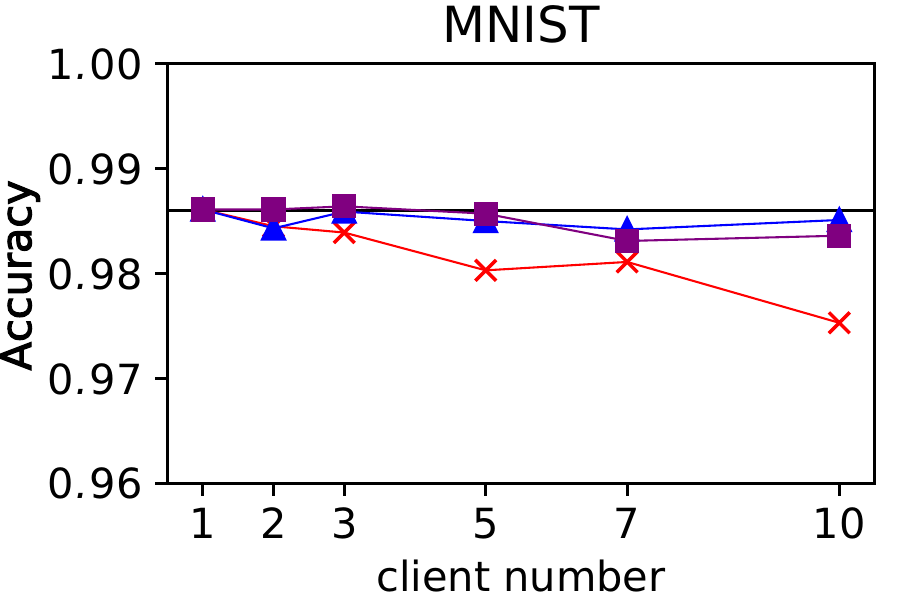}
  \includegraphics[width=0.215\textwidth]{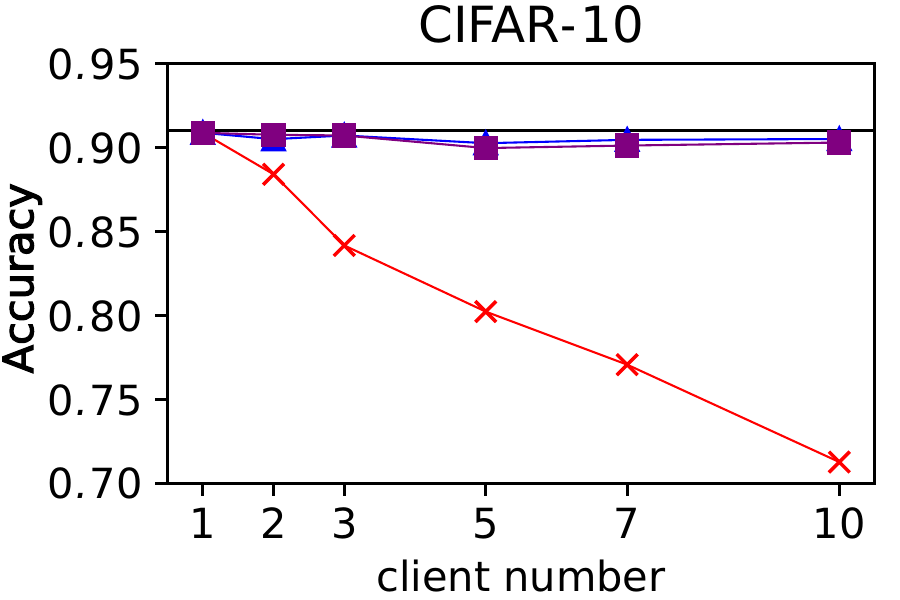}
  \label{fig:LC CN}
  }

  \subfigure[The accuracy against the parameter $\alpha$]{
  \includegraphics[width=0.215\textwidth]{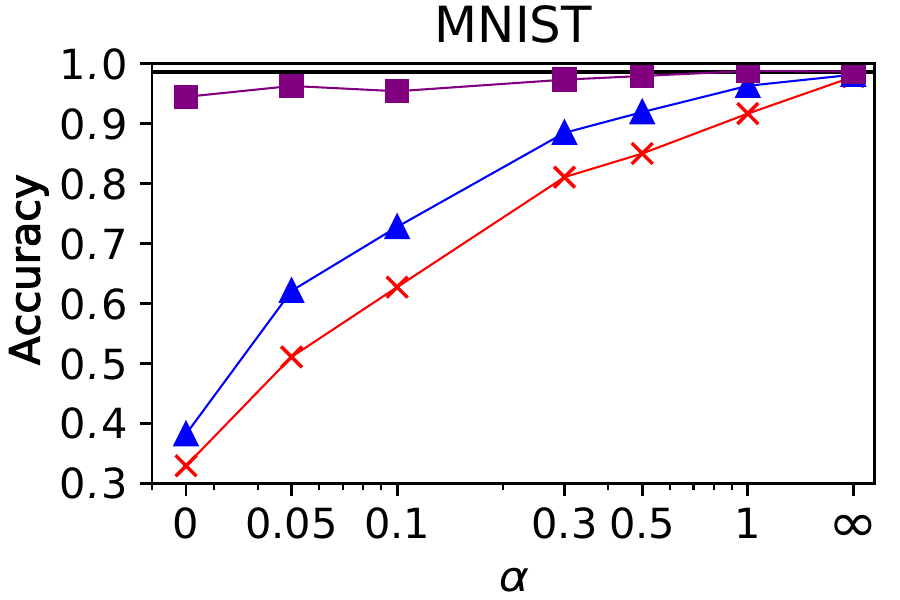}
  \includegraphics[width=0.215\textwidth]{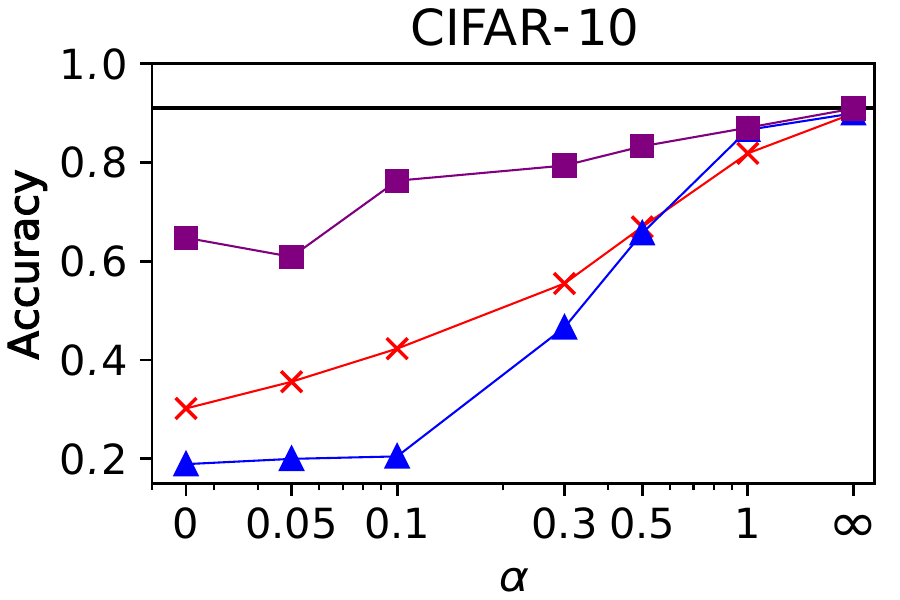}
  \label{fig:LC alpha}
  }
  
  \caption{The effect of client number and $\alpha$ on model accuracy.}
  \label{fig:LC}
  
\end{figure}
\nosection{IID settings}  The results of local training and two SL paradigms under IID settings are presented in Figure~\ref{fig:LC CN}. On the MNIST dataset, there is slight difference among the three algorithms, with only a small drop in local training accuracy at 10 clients. However, on the CIFAR-10 dataset, local training shows a significant drop, with an accuracy of only $71.47\%$ at 10 clients, which is probably due to the complexity of the learning task. 
Unlike handwritten characters, CIFAR-10 contains real-world objects, necessitating a sufficient amount of training data. Both SL paradigms have shown promising results, indicating that they can improve each client's model performance without exchanging the original data.

\nosection{Non-IID settings} The performance of local training and two SL paradigms under varying $\alpha$ in the Non-IID setting is presented in Figure~\ref{fig:LC alpha}. On the MNIST dataset, the performance of both Vanilla and local training decreases as $\alpha$ decreases, but the overall performance of Vanilla is still higher than local training. 
When $\alpha$ is 0, the accuracy of Vanilla also drops to a level similar to local training. On the CIFAR-10 dataset, Vanilla's performance is lower than expected, when $\alpha$ is less than 0.5, it has lower accuracy than local training. 

SplitFed performs well on both datasets, but on the CIFAR-10 dataset, there is a large drop in accuracy when $\alpha$ is below 0.1. At the most extreme Non-IID setting(i.e., $\alpha=0$), SplitFed achieves an accuracy of 64.73$\%$. SplitFed also has difficulty converging in this case, but overall its performance is still better than the other two paradigms. SplitFed can mitigate this problem by increasing the frequency of aggregation. 

In general, Vanilla can help clients improve performance when data is evenly distributed. However, when the task is complex and the Non-IID distribution is extreme, we recommend using SplitFed to help local devices improve performance.

%% file: ijcai23.bbl
\begin{thebibliography}{}

\bibitem[\protect\citeauthoryear{Abuadbba \bgroup \em et al.\egroup
  }{2020}]{DBLP:journals/corr/abs-2003-12365}
Sharif Abuadbba, Kyuyeon Kim, Minki Kim, Chandra Thapa, Seyit~Ahmet
  {\c{C}}amtepe, Yansong Gao, Hyoungshick Kim, and Surya Nepal.
\newblock Can we use split learning on 1d {CNN} models for privacy preserving
  training?
\newblock {\em CoRR}, abs/2003.12365, 2020.

\bibitem[\protect\citeauthoryear{Acevedo{-}Viloria \bgroup \em et al.\egroup
  }{2021}]{DBLP:journals/corr/abs-2107-13673}
Jaime~D. Acevedo{-}Viloria, Luisa Roa, Soji Adeshina, Cesar~Charalla Olazo,
  Andr{\'{e}}s Rodr{\'{\i}}guez{-}Rey, Jose~Alberto Ramos, and Alejandro~Correa
  Bahnsen.
\newblock Relational graph neural networks for fraud detection in a super-app
  environment.
\newblock {\em CoRR}, abs/2107.13673, 2021.

\bibitem[\protect\citeauthoryear{Alessandro and
  Mantelero}{2013}]{Alessandro2013The}
Alessandro and Mantelero.
\newblock The eu proposal for a general data protection regulation and the
  roots of the 'right to be forgotten'.
\newblock {\em Computer law \& security report}, 29(3):229--235, 2013.

\bibitem[\protect\citeauthoryear{Baek \bgroup \em et al.\egroup
  }{2022}]{2022Visual}
S.~Baek, J.~Park, P.~Vepakomma, R.~Raskar, M.~Bennis, and S.~L. Kim.
\newblock Visual transformer meets cutmix for improved accuracy, communication
  efficiency, and data privacy in split learning.
\newblock {\em arXiv e-prints}, 2022.

\bibitem[\protect\citeauthoryear{Beutel \bgroup \em et al.\egroup
  }{2020}]{FLOWER}
Daniel~J. Beutel, Taner Topal, Akhil Mathur, Xinchi Qiu, Titouan Parcollet, and
  Nicholas~D. Lane.
\newblock Flower: {A} friendly federated learning research framework.
\newblock {\em CoRR}, abs/2007.14390, 2020.

\bibitem[\protect\citeauthoryear{Caldas \bgroup \em et al.\egroup
  }{2018}]{DBLP:journals/corr/abs-1812-01097}
Sebastian Caldas, Peter Wu, Tian Li, Jakub Kone{\v{c}}n{\'y}, H.~Brendan
  McMahan, Virginia Smith, and Ameet Talwalkar.
\newblock {LEAF:} {A} benchmark for federated settings.
\newblock {\em CoRR}, abs/1812.01097, 2018.

\bibitem[\protect\citeauthoryear{Chen \bgroup \em et al.\egroup
  }{2021}]{2021Communication}
Xing Chen, Jingtao Li, and Chaitali Chakrabarti.
\newblock Communication and computation reduction for split learning using
  asynchronous training.
\newblock {\em CoRR}, abs/2107.09786, 2021.

\bibitem[\protect\citeauthoryear{Cheng-Yen~Hsieh}{2022}]{3C_SL}
An-Yeu~(Andy)Wu Cheng-Yen~Hsieh, Yu-Chuan~Chuang.
\newblock C3-sl: Circular convolution-based batch-wise compression for
  communication-efficient split learning.
\newblock {\em CoRR}, abs/2207.12397, 2022.

\bibitem[\protect\citeauthoryear{Chervenak \bgroup \em et al.\egroup
  }{2000}]{CHERVENAK2000187}
Ann Chervenak, Ian Foster, Carl Kesselman, Charles Salisbury, and Steven
  Tuecke.
\newblock The data grid: Towards an architecture for the distributed management
  and analysis of large scientific datasets.
\newblock {\em Journal of Network and Computer Applications}, 23(3):187--200,
  2000.

\bibitem[\protect\citeauthoryear{Cun \bgroup \em et al.\egroup
  }{1990}]{1990Handwritten}
Y.~L. Cun, B.~Boser, J.~S. Denker, D.~Henderson, and L.~D. Jackel.
\newblock Handwritten digit recognition with a back-propagation network.
\newblock {\em Advances in neural information processing systems},
  2(2):396--404, 1990.

\bibitem[\protect\citeauthoryear{Goyal \bgroup \em et al.\egroup
  }{2017}]{DBLP:journals/corr/GoyalDGNWKTJH17}
Priya Goyal, Piotr Doll{\'{a}}r, Ross~B. Girshick, Pieter Noordhuis, Lukasz
  Wesolowski, Aapo Kyrola, Andrew Tulloch, Yangqing Jia, and Kaiming He.
\newblock Accurate, large minibatch {SGD:} training imagenet in 1 hour.
\newblock {\em CoRR}, abs/1706.02677, 2017.

\bibitem[\protect\citeauthoryear{Gupta and
  Raskar}{2018}]{DBLP:journals/corr/abs-1810-06060}
Otkrist Gupta and Ramesh Raskar.
\newblock Distributed learning of deep neural network over multiple agents.
\newblock {\em CoRR}, abs/1810.06060, 2018.

\bibitem[\protect\citeauthoryear{He \bgroup \em et al.\egroup }{2016}]{ResNet}
Kaiming He, Xiangyu Zhang, Shaoqing Ren, and Jian Sun.
\newblock Deep residual learning for image recognition.
\newblock In {\em 2016 IEEE Conference on Computer Vision and Pattern
  Recognition (CVPR)}, pages 770--778, 2016.

\bibitem[\protect\citeauthoryear{He \bgroup \em et al.\egroup
  }{2020}]{DBLP:journals/corr/abs-2007-13518}
Chaoyang He, Songze Li, Jinhyun So, Mi~Zhang, Hongyi Wang, Xiaoyang Wang,
  Praneeth Vepakomma, Abhishek Singh, Hang Qiu, Li~Shen, Peilin Zhao, Yan Kang,
  Yang Liu, Ramesh Raskar, Qiang Yang, Murali Annavaram, and Salman Avestimehr.
\newblock Fedml: {A} research library and benchmark for federated machine
  learning.
\newblock {\em CoRR}, abs/2007.13518, 2020.

\bibitem[\protect\citeauthoryear{Hsieh \bgroup \em et al.\egroup
  }{2022}]{DBLP:journals/corr/abs-2207-12397}
Cheng{-}Yen Hsieh, Yu{-}Chuan Chuang, and An{-}Yeu Wu.
\newblock {C3-SL:} circular convolution-based batch-wise compression for
  communication-efficient split learning.
\newblock {\em CoRR}, abs/2207.12397, 2022.

\bibitem[\protect\citeauthoryear{Huang \bgroup \em et al.\egroup
  }{2019}]{HUANG2019103291}
Li~Huang, Andrew~L. Shea, Huining Qian, Aditya Masurkar, Hao Deng, and Dianbo
  Liu.
\newblock Patient clustering improves efficiency of federated machine learning
  to predict mortality and hospital stay time using distributed electronic
  medical records.
\newblock {\em Journal of Biomedical Informatics}, 99:103291, 2019.

\bibitem[\protect\citeauthoryear{Jeon and Kim}{2020}]{2020Privacy}
J.~Jeon and J.~Kim.
\newblock Privacy-sensitive parallel split learning.
\newblock In {\em 2020 International Conference on Information Networking
  (ICOIN)}, 2020.

\bibitem[\protect\citeauthoryear{Kohavi}{1997}]{osti_421279}
R.~Kohavi.
\newblock Scaling up the accuracy of naive-bayes classifiers: a decision-tree
  hybrid.
\newblock {\em Proc of Kdd}, 1997.

\bibitem[\protect\citeauthoryear{Kone{\v{c}}n{\'y} \bgroup \em et al.\egroup
  }{2016}]{LSDG2}
Jakub Kone{\v{c}}n{\'y}, H.~Brendan McMahan, Felix~X. Yu, Peter
  Richt{\'{a}}rik, Ananda~Theertha Suresh, and Dave Bacon.
\newblock Federated learning: Strategies for improving communication
  efficiency.
\newblock {\em CoRR}, abs/1610.05492, 2016.

\bibitem[\protect\citeauthoryear{Krizhevsky \bgroup \em et al.\egroup
  }{2017}]{Krizhevsky2012ImageNet}
Alex Krizhevsky, Ilya Sutskever, and Geoffrey~E. Hinton.
\newblock Imagenet classification with deep convolutional neural networks.
\newblock {\em Commun. ACM}, 60(6):84–90, may 2017.

\bibitem[\protect\citeauthoryear{Liu \bgroup \em et al.\egroup
  }{2017}]{DBLP:journals/corr/LiuGNDKBVTNCHPS17}
Yun Liu, Krishna Gadepalli, Mohammad Norouzi, George~E. Dahl, Timo Kohlberger,
  Aleksey Boyko, Subhashini Venugopalan, Aleksei Timofeev, Philip~Q. Nelson,
  Gregory~S. Corrado, Jason~D. Hipp, Lily Peng, and Martin~C. Stumpe.
\newblock Detecting cancer metastases on gigapixel pathology images.
\newblock {\em CoRR}, abs/1703.02442, 2017.

\bibitem[\protect\citeauthoryear{Liu \bgroup \em et al.\egroup
  }{2022}]{857534b2836744119e61880381fa08e8_HFSL}
Xiaolan Liu, Yansha Deng, and Toktam Mahmoodi.
\newblock {\em Energy Efficient User Scheduling for Hybrid Split and Federated
  Learning in Wireless UAV Networks}.
\newblock IEEE Communications Society, United States, 2022.

\bibitem[\protect\citeauthoryear{McMahan \bgroup \em et al.\egroup
  }{2016}]{2016Communication}
H.~Brendan McMahan, Eider Moore, Daniel Ramage, and Blaise~Ag{\"{u}}era
  y~Arcas.
\newblock Federated learning of deep networks using model averaging.
\newblock {\em CoRR}, abs/1602.05629, 2016.

\bibitem[\protect\citeauthoryear{Oh \bgroup \em et al.\egroup
  }{2022}]{10.1145/3485447.3512153}
Seungeun Oh, Jihong Park, Praneeth Vepakomma, Sihun Baek, Ramesh Raskar, Mehdi
  Bennis, and Seong-Lyun Kim.
\newblock Locfedmix-sl: Localize, federate, and mix for improved scalability,
  convergence, and latency in split learning.
\newblock In {\em Proceedings of the ACM Web Conference 2022}, WWW '22, page
  3347–3357, New York, NY, USA, 2022. Association for Computing Machinery.

\bibitem[\protect\citeauthoryear{Pal \bgroup \em et al.\egroup
  }{2021}]{2021Server}
Shraman Pal, Mansi Uniyal, Jihong Park, Praneeth Vepakomma, Ramesh Raskar,
  Mehdi Bennis, Moongu Jeon, and Jinho Choi.
\newblock Server-side local gradient averaging and learning rate acceleration
  for scalable split learning.
\newblock {\em CoRR}, abs/2112.05929, 2021.

\bibitem[\protect\citeauthoryear{Park \bgroup \em et al.\egroup }{2021}]{FeSVT}
Sangjoon Park, Gwanghyun Kim, Jeongsol Kim, Boah Kim, and Jong~Chul Ye.
\newblock Federated split vision transformer for covid-19cxr diagnosis using
  task-agnostic training, 2021.

\bibitem[\protect\citeauthoryear{Plate and T.}{1995}]{Plate1995Holographic}
Plate and A.~T.
\newblock Holographic reduced representations.
\newblock {\em Neural Networks, IEEE Transactions on}, 1995.

\bibitem[\protect\citeauthoryear{Poirot \bgroup \em et al.\egroup
  }{2019}]{DBLP:journals/corr/abs-1912-12115}
Maarten~G. Poirot, Praneeth Vepakomma, Ken Chang, Jayashree Kalpathy{-}Cramer,
  Rajiv Gupta, and Ramesh Raskar.
\newblock Split learning for collaborative deep learning in healthcare.
\newblock {\em CoRR}, abs/1912.12115, 2019.

\bibitem[\protect\citeauthoryear{Ryffel \bgroup \em et al.\egroup
  }{2018}]{DBLP:journals/corr/abs-1811-04017}
Th{\'{e}}o Ryffel, Andrew Trask, Morten Dahl, Bobby Wagner, Jason Mancuso,
  Daniel Rueckert, and Jonathan Passerat{-}Palmbach.
\newblock A generic framework for privacy preserving deep learning.
\newblock {\em CoRR}, abs/1811.04017, 2018.

\bibitem[\protect\citeauthoryear{Shickel \bgroup \em et al.\egroup
  }{2018}]{8086133}
Benjamin Shickel, Patrick~James Tighe, Azra Bihorac, and Parisa Rashidi.
\newblock Deep ehr: A survey of recent advances in deep learning techniques for
  electronic health record (ehr) analysis.
\newblock {\em IEEE Journal of Biomedical and Health Informatics},
  22(5):1589--1604, 2018.

\bibitem[\protect\citeauthoryear{Thapa \bgroup \em et al.\egroup
  }{2020}]{SplitFed}
Chandra Thapa, Mahawaga Arachchige~Pathum Chamikara, and Seyit Camtepe.
\newblock Splitfed: When federated learning meets split learning.
\newblock {\em CoRR}, abs/2004.12088, 2020.

\bibitem[\protect\citeauthoryear{Turina \bgroup \em et al.\egroup
  }{2021}]{9582171}
Valeria Turina, Zongshun Zhang, Flavio Esposito, and Ibrahim Matta.
\newblock Federated or split? a performance and privacy analysis of hybrid
  split and federated learning architectures.
\newblock In {\em 2021 IEEE 14th International Conference on Cloud Computing
  (CLOUD)}, pages 250--260, 2021.

\bibitem[\protect\citeauthoryear{Vepakomma \bgroup \em et al.\egroup
  }{2018}]{DBLP:journals/corr/abs-1812-03288}
Praneeth Vepakomma, Tristan Swedish, Ramesh Raskar, Otkrist Gupta, and
  Abhimanyu Dubey.
\newblock No peek: {A} survey of private distributed deep learning.
\newblock {\em CoRR}, abs/1812.03288, 2018.

\bibitem[\protect\citeauthoryear{Wang \bgroup \em et al.\egroup
  }{2016}]{DBLP:journals/corr/WangGZXGL16}
Qinglong Wang, Wenbo Guo, Kaixuan Zhang, Xinyu Xing, C.~Lee Giles, and Xue Liu.
\newblock Random feature nullification for adversary resistant deep
  architecture.
\newblock {\em CoRR}, abs/1610.01239, 2016.

\bibitem[\protect\citeauthoryear{Wu \bgroup \em et al.\egroup
  }{2022}]{2022Split_CPSL}
W.~Wu, M.~Li, K.~Qu, C.~Zhou, Xuemin, Shen, W.~Zhuang, X.~Li, and W.~Shi.
\newblock Split learning over wireless networks: Parallel design and resource
  management.
\newblock {\em arXiv e-prints}, 2022.

\end{thebibliography}
